\documentclass[runningheads]{llncs}
\usepackage{hyperref}
\usepackage{graphicx}
\usepackage{subfig}
\usepackage{array,multirow}
\usepackage{amsmath,amssymb} 
\usepackage{color}
\usepackage[labelfont=bf]{caption}
\usepackage{tabularx}
\usepackage{booktabs} 
\usepackage{url}

\begin{document}

\newcommand\T{\rule{0pt}{2.2ex}}       
\newcommand\B{\rule[-0.8ex]{0pt}{0pt}} 
\def\etal{\emph{et al}.}

\title{Diagnostics in Semantic Segmentation} 

\titlerunning{Diagnostics in Semantic Segmentation}


\author{Vladimir Nekrasov \and
Chunhua Shen \and
Ian Reid}

%





\authorrunning{V. Nekrasov, C. Shen, I. Reid}

%
\institute{School of Computer Science, University of Adelaide, Australia \
	\email{\{firstname.lastname\}@adelaide.edu.au}}

\maketitle

\begin{abstract}
Over the past years, computer vision community has contributed to enormous progress in semantic image segmentation, a per-pixel classification task, crucial for dense scene understanding and rapidly becoming vital in lots of real-world applications, including driverless cars and medical imaging. Most recent models are now reaching previously unthinkable numbers (e.g., $89\%$ mean iou on PASCAL VOC, $83\%$ on CityScapes), and, while intersection-over-union and a range of other metrics provide the general picture of model performance, in this paper we aim to extend them into other meaningful and important for applications characteristics, answering such questions as `how accurate the model segmentation is on small objects in the general scene?', or `what are the sources of uncertainty that cause the model to make an erroneous prediction?'. Besides establishing a methodology that covers the performance of a single model from different perspectives, we also showcase several extensions that can be worth pursuing in order to further improve current results in semantic segmentation.
\end{abstract}

\section{Introduction}

Most practical systems must be evaluated on all sorts of benchmarks, and a multitude of different metrics must be computed in order to make the decision upon acceptance of the system as functional or faulty. The same applies to deep learning models, competing against each other on carefully chosen benchmarks. Nevertheless, none of those common benchmarks for deep learning, in general, and semantic segmentation, in particular, consider to go deeper into the numbers and look at the given problem from a different angle. For example, none of them will provide you with an understanding whether your car detector completely fails at recognising all cars, or only small cars, or only cars located near buses. Given suitable data, it is possible to answer such questions, and here we show the value of doing that by highlighting the failure modes of the chosen networks. Failure modes are an essential part of our (human) learning process, and thus it motivates us to analyse failure modes of modern semantic segmentation models in a more detailed way.

We have chosen semantic segmentation as it is a critical component of scene understanding, and already finds its niche in many sorts of applications, ranging from driverless cars~\cite{XuGYD17} to medical imaging~\cite{RonnebergerFB15}. Besides that, there already exists several excellent diagnostics works in other domains~\cite{HoiemCD12,HariharanAGM14,RonchiP17}, which motivates us to extend them for semantic segmentation. Our aim here is to encourage researchers and practitioners to look at semantic segmentation performance from all sorts of different angles, ranging from the connections between object size and per-pixel accuracy to different notions of uncertainty and error taxonomy. 

In particular, we consider two state-of-the-art models on two standard datasets for semantic segmentation, namely, PASCAL VOC~\cite{EveringhamGWWZ10}, and CityScapes~\cite{CordtsORREBFRS16}, where performance levels on generic benchmarks, such as intersection-over-union and per-pixel accuracy, have already reached a very high bar. Besides that, both of them provide per-instance annotations, which gives us an opportunity to reason about model performance in terms of object characteristics. Each model on each dataset we describe with regard to its sensitivity to object size and aspect ratio, error taxonomy, uncertainty levels and their correlation with performance. Finally, for each case discussed, we showcase simple extensions and provide our recommendations about possible future research directions.

Our methodology is general and flexible, making it straightforward to retrieve the same characteristics for all sorts of models and datasets. 

\section{Related Work}

\subsection{Semantic Segmentation}
Semantic segmentation is the task where one is asked to predict a semantic label per each pixel in the image. Although similar in nature to image classification, it comprises several difficulties - one of which is dealing with variable input and output image sizes. First approaches employed sliding window methodology on fixed-size inputs~\cite{SchroffCZ08,YinBCK10} until Long~\etal~\cite{LongSD15} proposed a fully convolutional variant of image classifiers. This became the standard choice of solving any per-pixel tasks, and semantic segmentation, in particular, has witnessed a significant progress partially due to the development of end-to-end probabilistic graphical models~\cite{0001JRVSDHT15,LinSHR16}, and partially due to the advances in different structures and contextual modules~\cite{WuSH16e,ZhaoSQWJ17,abs-1802-02611}.

Here, we consider two state-of-the-art networks, DeepLab-v3~\cite{abs-1802-02611}, and ResNet-38~\cite{WuSH16e}, on two popular benchmarks, PASCAL VOC~\cite{EveringhamGWWZ10}, suitable for general segmentation, and CityScapes~\cite{CordtsORREBFRS16}, for more specific driving applications. DeepLab-v3 is a successor of the original DeepLab network~\cite{ChenPK0Y16} with the inclusion of encoder and a contextual module, containing dilated convolutions. ResNet-38 was proposed as an alternative to the original residual networks~\cite{HeZRS162} after a careful analysis conducted to evaluate the trade-off between depth and width of a convolutional network.\par
The authors provide the models not pre-trained on the validation sets, on which these networks show comparable results, allowing us to make a fair comparison. 
We should also note that in this work we do not aim to attribute a particular rise in performance to any of structural and architectural advances, rather our goal is to underline similarities, strong and weak spots across different models.
\subsection{Diagnostics of computer vision methods}
The role of diagnostics in computer vision has often been overlooked in recent years. Nevertheless, some pivotal works cannot go unnoticed. 

For object detection, Hoiem~\etal~\cite{HoiemCD12} 
compared two state-of-the-art models by analysing their false positives and false negatives on the PASCAL VOC dataset~\cite{EveringhamGWWZ10}. The analysis was based on different properties of the object instances, such as occlusion, truncation, visibility and size. From it, the authors were able to pinpoint strengths and weaknesses of the methods, in particular, the sensitivity to large and small objects, and different levels of occlusion. We are primarily motivated by this work, and aim to extend it for semantic segmentation, but we consider per-instance annotations only as other properties are not well-annotated.

Later, Hariharan~\etal~\cite{HariharanAGM14} conducted a similar analysis for object detection by analysing error modes and sources of false positives. They concluded that mislocalisation was the single most influential source of errors for object detectors.

Most recently, Ronchi \& Perona~\cite{RonchiP17} built a diagnostics framework for multi-instance pose estimation. Specifically, they proposed a taxonomy of false positive localisation errors, which includes \emph{`miss'}, \emph{`swap'}, \emph{`jitter'} and \emph{`inversion'}. Based on the proposed taxonomy, they evaluated two state-of-the-art models and underlined which error modes were of the most influence. They concluded that, besides missing keypoints, those models were also suffering from noise in confidence scores, which negatively affected their performance.

On a related note, there have been multiple attempts proposing the most suitable set of evaluation metrics for semantic segmentation~\cite{FreixenetMRMC02,UnnikrishnanPH05,ShottonWRC06}. For example, Csurka~\etal~\cite{CsurkaLP13} argue that dataset-level metrics are less meaningful than image-level ones, as the latter allow to statistically quantify differences between two methods and better analyse their performance. We partially follow this approach and, besides reporting global per-pixel accuracy and intersection-over-union, we also report per-instance accuracy, which enables us to reason about performance across different instance properties, such as size and aspect ratio. 

\section{Methodology}


\begin{figure}[htb]
\centering
\resizebox{0.7\textwidth}{!}{
\begin{tabular}{cccccc}
  & XS & S & M & L & XL\\[-0.15in]
  \raisebox{6.\height}{XT} & \subfloat{\includegraphics[width = 0.19\linewidth]{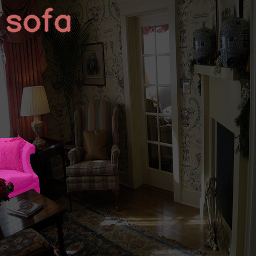}}
     & \subfloat{\includegraphics[width = 0.19\linewidth]{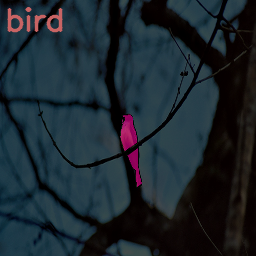}}
     & \subfloat{\includegraphics[width = 0.19\linewidth]{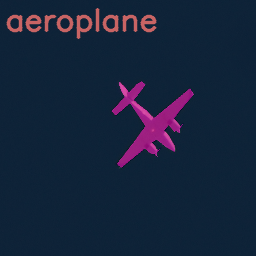}}
     & \subfloat{\includegraphics[width = 0.19\linewidth]{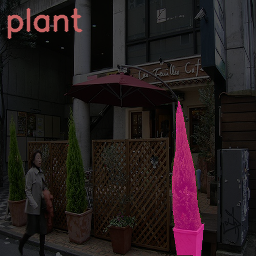}}
     & \subfloat{\includegraphics[width = 0.19\linewidth]{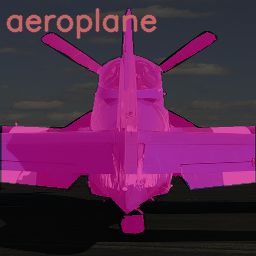}}\\[-0.15in]
  \raisebox{6.\height}{T} & \subfloat{\includegraphics[width = 0.19\linewidth]{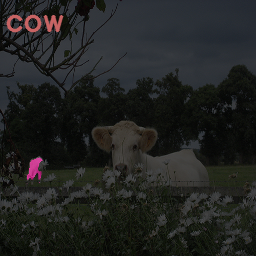}}
     & \subfloat{\includegraphics[width = 0.19\linewidth]{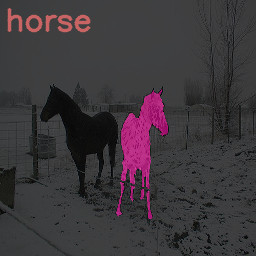}}
     & \subfloat{\includegraphics[width = 0.19\linewidth]{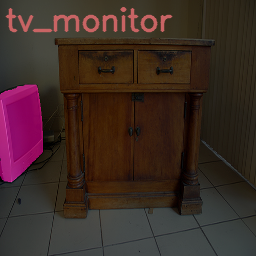}}
     & \subfloat{\includegraphics[width = 0.19\linewidth]{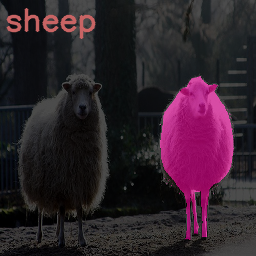}}
     & \subfloat{\includegraphics[width = 0.19\linewidth]{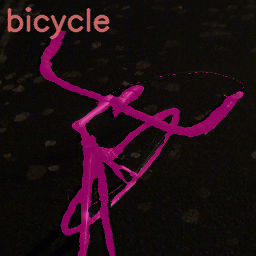}}\\[-0.15in]
  \raisebox{6.\height}{M} & \subfloat{\includegraphics[width = 0.19\linewidth]{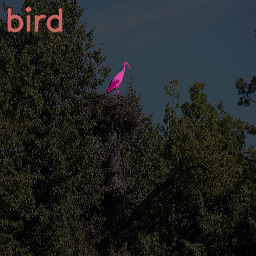}}
     & \subfloat{\includegraphics[width = 0.19\linewidth]{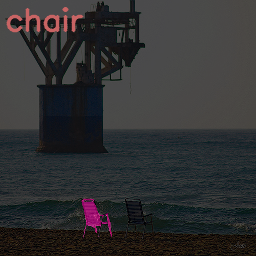}}
     & \subfloat{\includegraphics[width = 0.19\linewidth]{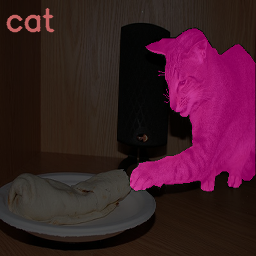}}
     & \subfloat{\includegraphics[width = 0.19\linewidth]{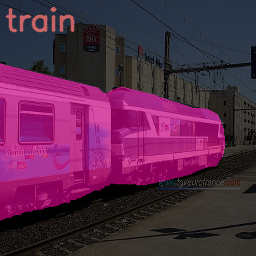}}
     & \subfloat{\includegraphics[width = 0.19\linewidth]{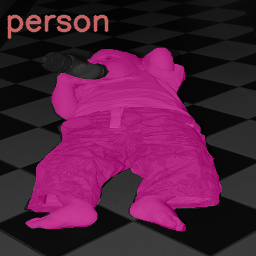}}\\[-0.15in]
  \raisebox{6.\height}{W} & \subfloat{\includegraphics[width = 0.19\linewidth]{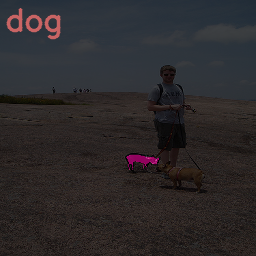}}
     & \subfloat{\includegraphics[width = 0.19\linewidth]{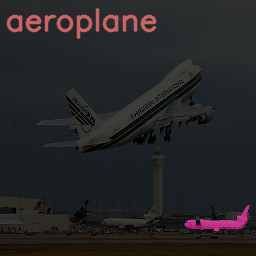}}
     & \subfloat{\includegraphics[width = 0.19\linewidth]{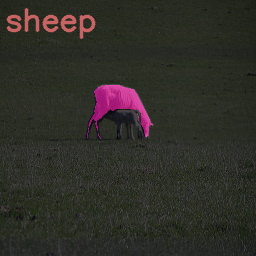}}
     & \subfloat{\includegraphics[width = 0.19\linewidth]{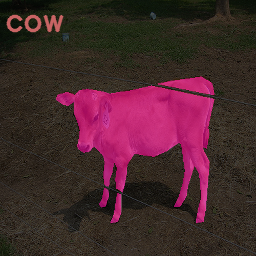}}
     & \subfloat{\includegraphics[width = 0.19\linewidth]{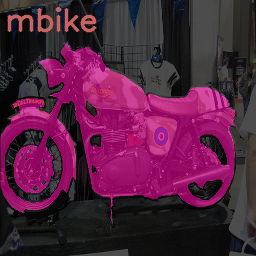}}\\[-0.15in]
  \raisebox{6.\height}{XW} & \subfloat{\includegraphics[width = 0.19\linewidth]{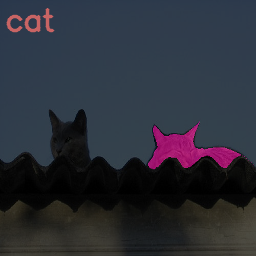}}
     & \subfloat{\includegraphics[width = 0.19\linewidth]{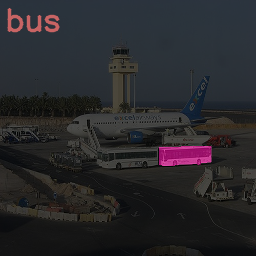}}
     & \subfloat{\includegraphics[width = 0.19\linewidth]{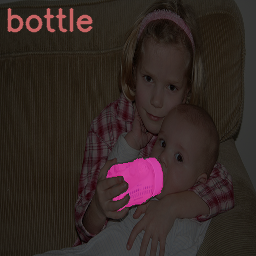}}
     & \subfloat{\includegraphics[width = 0.19\linewidth]{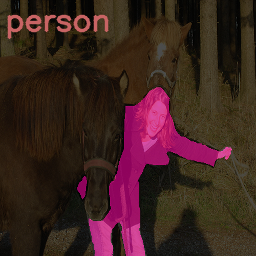}}
     & \subfloat{\includegraphics[width = 0.19\linewidth]{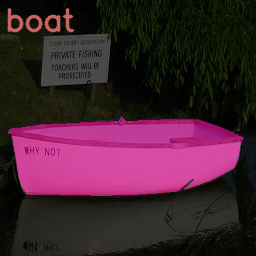}}
\end{tabular}
}
\caption{Examples of instances of different sizes (horisontal axis) and aspect ratios (vertical axis) on the validation set of PASCAL VOC. The relevant instance on each image is highlighted in pink, and its semantic class is given in the top left corner of the image\label{fig:voc-val-ex}}
\end{figure}

\subsection{Object Characteristics}
As noted by both Hoiem~\etal~\cite{HoiemCD12} and Harihan~\etal~\cite{HariharanAGM14}, object characteristics tend to have a large impact on the model performance. Motivated by this, we first consider how sensitive each network is to size and aspect ratio of object instances. As we have access to instance annotations, we follow Hoiem~\etal~\cite{HoiemCD12} and divide the instances of each class into one of five size categories based on the number of pixels that each instance has - extra-small (XS: bottom $10\%$), small (S: $10$-$30\%$), medium (M: $30$-$70\%$), large (L: $70$-$90\%$) and extra-large (XL: top $10\%$). Analogously, we divide instances into five aspect ratio ($\frac{width}{height}$) categories - extra-tall (XT), tall (T), medium (M), wide (W), extra-wide (XW). When considered together, these definitions cover a broad spectrum of different instances - from sparse but spacious to dense but tiny. We demonstrate examples on Fig.~\ref{fig:voc-val-ex}.\par
Using such groupings, we are able to quantify model performance in terms of size and aspect ratio of each class instance.

\subsection{Error Taxonomy}
We move on to describe different sources of errors, namely, \textit{mislocalisation} - when the predicted label is incorrect, but it does exist in the ground truth mask in close vicinity (here we consider a square patch centered at the prediction point); and \textit{confusion} with other labels. Confusion can be of three different types: confusion with semantically similar classes - when the predicted label shares a subclass with the ground truth label; confusion with background (in case such a label exists) - when the predicted label is background, but the ground truth one is not; and confusion with semantically dissimilar classes - in all other cases. For grouping of semantically similar classes on PASCAL VOC, we follow Hoiem~\etal~\cite{HoiemCD12}, whereas for CityScapes we make use of the provided hierarchy (Table~\ref{table:char}).\par
Such an error taxonomy 
 enables us to evaluate the impact of each type of error separately, potentially leading to a specific algorithm built to alleviate the effect of each group.

\setlength{\tabcolsep}{4pt}
\begin{table}[htb]
	\begin{center}
	\caption{Semantically similar classes in PASCAL VOC and CityScapes
		\label{table:char}}
		\begin{tabular}{c|c|c}
			\specialrule{.15em}{0em}{0em} 
			Dataset & Grouping & Classes\T\B\\
			\specialrule{.1em}{0em}{0em}
			\hline
			\parbox[t]{2mm}{\multirow{4}{*}{\rotatebox[origin=c]{90}{VOC}}} 	& Aero & aeroplane, bird\T\\
			&Animals+Human & cat, cow, dog, horse, person, sheep\T\\
			&Furniture & chair, sofa, table\T\\
			&Vehicles & bicycle, boat, bus, car, mbike, train\B\\
			\specialrule{.15em}{0em}{0em}
			\hline
			\parbox[t]{2mm}{\multirow{6}{*}{\rotatebox[origin=c]{90}{CityScapes}}} 	& Construction & building, wall, fence\T\\
			&Flat & road, sidewalk\T\\
			&Human & person, rider\T\\
			&Nature & vegetation, terrain\T\\
			&Object & pole, traffic light, traffic sign\T\\
			&Vehicles & car, truck, bus, train, motorcycle, bicycle\B\\
			\specialrule{.15em}{0em}{0em}
		\end{tabular}
	\end{center}
	\vskip -0.35in
\end{table}
\setlength{\tabcolsep}{1.4pt}

\subsection{Quantifying Uncertainty}
Uncertainty is an important part of any functional system, and knowing sources of it might shed light on system's behaviour.\par
Here, we exploit the softmax approximation to acquire per-pixel probabilities of each class from the model's outputs, and, based on it, define two notions of uncertainty - per-pixel relative entropy and relative probability difference between top-$2$ and top-$1$ predicted classes:

\begin{align}
\label{eqn1}
\begin{gathered}
Relative~Entropy\stackrel{\text{def}}{=}\frac{\sum_{c\in{C}}{p_{c} \cdot log(p_{c})}}{log(\frac{1}{|C|})},\\
Relative~Probability\stackrel{\text{def}}{=}\frac{p_{top-2}}{p_{top-1}},
\end{gathered}
\end{align}
where $C$ is the set of semantic classes, and $p_{c}$ is the predicted probability of class $c$ at the given pixel. Both measures range from $0$ to $1$, where the higher values denote the higher uncertainty of the model in its own predictions. By tying up such means of uncertainty with the error types and object characteristics defined above, we are able to answer the following sorts of questions: if the instance is undetected (confused with background), how much does its uncertainty deviates from the average one? Or how does uncertainty differ across objects of various sizes?



\section{Results}

We consider two datasets - PASCAL VOC~\cite{EveringhamGWWZ10} and CityScapes~\cite{CordtsORREBFRS16}, and two networks - DeepLab-v3\footnote{\url{https://github.com/tensorflow/models/tree/master/research/deeplab}}~\cite{abs-1802-02611} and ResNet-38\footnote{\url{https://github.com/itijyou/ademxapp}}~\cite{WuSH16e}. PASCAL VOC contains a wide spectrum of $20$ semantic classes (plus additional `background' label), and has $1449$ images for validation, whereas CityScapes includes $500$ validation images annotated with $19$ semantic classes. While VOC provides instance-level annotations for all the classes, CityScapes has them present only for $8$ classes - `bicycle', `bus', `car', `motorcycle', `person', `train', `truck' and `rider'. We do not alter or fine-tune the provided weights in any way, and only amend the pre-processing steps to not include any rescaling of the input image and the post-processing step to only include bicubic upsampling of the score maps to the original size. In the interests of brevity, for each dataset and each defined terminology, we only discuss most interesting results on a subset of classes; we provide results for all the classes in our supplementary material\footnote{\url{https://cv-conf.shinyapps.io/diag-sem-segm/}}.\par
We report two well-established types of quantitative measures for semantic segmentation - pixel accuracy ($\frac{s_{ii}}{g_{i}}$) and intersection-over-union ($\frac{s_{ii}}{g_{i}+\sum_{j\in{C}}{s_{ij}}-s_{ii}}$), where $s_{ij}$ is the number of pixels belonging to class $i$ while being predicted as class $j$ and $g_{i}$ is the total number of pixels belonging to class $i$. When possible, we report average pixel accuracy across instances (as opposed to global values across the whole validation set). Results of the models across the validation sets are given in Table~\ref{table:res-voc} for VOC and in Table~\ref{table:res-cs} for CityScapes, respectively.

\vskip -0.25in
\setlength{\tabcolsep}{4pt}
\begin{table}[htb]
	\begin{center}
		\caption{Results on the validation set of PASCAL VOC without any scaling during pre-preprocessing, at non-background pixels in the ground truth maps
			\label{table:res-voc}}
		\resizebox{\textwidth}{!}{
		\begin{tabular}{c*{22}{|c}}
			\specialrule{.15em}{0em}{0em} 
			Metric & Model & aero & bike & bird & boat & bottle & bus & car & cat & chair & cow & table & dog & horse & mbike & person & plant & sheep & sofa & train & tv-mon & Total\T\B\\
			\specialrule{.1em}{0em}{0em}
			\hline
			\parbox[t]{1mm}{\multirow{2}{*}{\rotatebox[origin=c]{90}{Acc}}} & DeepLab-v3 & 97.83 & 93.23 & 95.44 & 91.49 & 90.95 & 98.06 & 95.50 & 98.07 & 69.96 & 97.37 & 62.77 & 96.09 & 96.97 & 94.23 & 93.88 & 76.23 & 96.25 & 67.51 & 98.18 & 83.28 & 89.66\T\\
			& ResNet-38 & 94.78 & 84.11 & 93.60 & 83.20 & 80.85 & 97.31 & 92.55 & 96.67 & 59.41 & 94.54 & 70.06 & 92.33 & 93.99 & 92.36 & 94.16 & 72.18 & 94.52 & 73.36 & 92.41 & 80.62 & 86.65\T\\
			\specialrule{.1em}{0em}{0em}
			\hline
			\parbox[t]{1mm}{\multirow{2}{*}{\rotatebox[origin=c]{90}{IoU}}} & DeepLab-v3 & 95.65 & 90.33 & 95.35 & 88.73 & 90.06 & 97.89 & 93.20 & 96.72 & 59.79 & 97.06 & 62.08 & 94.12 & 96.10 & 92.83 & 92.92 & 75.53 & 96.07 & 64.50 & 98.04 & 83.26 & 88.01\T\\
			& ResNet-38 &  94.38 & 82.05 & 93.36 & 82.76 & 80.08 & 97.09 & 92.11 & 94.85 & 53.97 & 92.97 & 69.08 & 89.70 & 93.09 & 90.93 & 92.70 & 71.28 & 93.39 & 67.11 & 92.32 & 80.55 & 85.19\B\\
			\specialrule{.15em}{0em}{0em}
		\end{tabular}
			}

	\end{center}
	\vskip -0.75in
\end{table}
\setlength{\tabcolsep}{1.4pt}

\setlength{\tabcolsep}{4pt}
\begin{table}[htb]
	\begin{center}
		\caption{Results on the validation set of CityScapes
			\label{table:res-cs}}
		\resizebox{\textwidth}{!}{
			\begin{tabular}{c*{21}{|c}}
			\specialrule{.15em}{0em}{0em} 
			Metric & Model & road & swalk & bldg & wall & fence & pole & t.light & t.sign & veget. & terrain & sky & person & rider & car & truck & bus & train & mcycle & bike & Total\T\B\\
			\specialrule{.1em}{0em}{0em}
			\hline
			\parbox[t]{1mm}{\multirow{2}{*}{\rotatebox[origin=c]{90}{Acc}}} & DeepLab-v3 & 98.86 & 93.79 & 96.60 & 64.94 & 71.05 & 80.76 & 81.63 & 86.82 & 96.68 & 73.19 & 98.11 & 91.64 & 75.48 & 97.72 & 89.84 & 93.57 & 84.57 & 77.70 & 88.35 & 86.38\T\\
			& ResNet-38 & 98.77 & 92.99 & 96.68 & 65.12 & 71.11 & 72.18 & 83.17 & 85.75 & 96.58 & 73.89 & 97.43 & 91.70 & 77.87 & 97.83 & 65.53 & 93.38 & 84.25 & 79.38 & 88.56 & 84.85\T\\
			\specialrule{.1em}{0em}{0em}
			\hline
			\parbox[t]{1mm}{\multirow{2}{*}{\rotatebox[origin=c]{90}{IoU}}} & DeepLab-v3 & 98.20 & 85.18 & 92.80 & 57.85 & 62.62 & 66.15 & 70.00 & 79.75 & 92.75 & 63.50 & 95.40 & 82.66 & 63.10 & 95.48 & 85.44 & 89.31 & 80.83 & 65.63 & 77.71 & 79.18\T\\
			& ResNet-38 &  97.93 & 83.87 & 92.55 & 58.37 & 61.19 & 62.35 & 70.49 & 78.69 & 92.33 & 63.73 & 94.12 & 82.92 & 65.18 & 94.64 & 60.88 & 88.37 & 81.40 & 68.82 & 77.93 & 77.67\B\\
			\specialrule{.15em}{0em}{0em}
			\end{tabular}
		}

	\end{center}
	\vskip -0.5in
\end{table}
\setlength{\tabcolsep}{1.4pt}

\subsection{Object Characteristics}

We examine classes `bottle', `car', `person' and `sofa' (VOC), and `motorcycle', `rider', `train' and `truck' (CityScapes), as these classes illustrate well the behaviour of model performance in terms of object characteristics among all the classes.
\subsubsection{Observations}
\begin{figure}[htb]
	\subfloat[PASCAL VOC - Size\label{fig:voc-ar-size-sen}]{%
		\begin{minipage}{0.5\linewidth}
			\centering
			\includegraphics[width = 1.\linewidth, trim=4 20 4 4,clip]{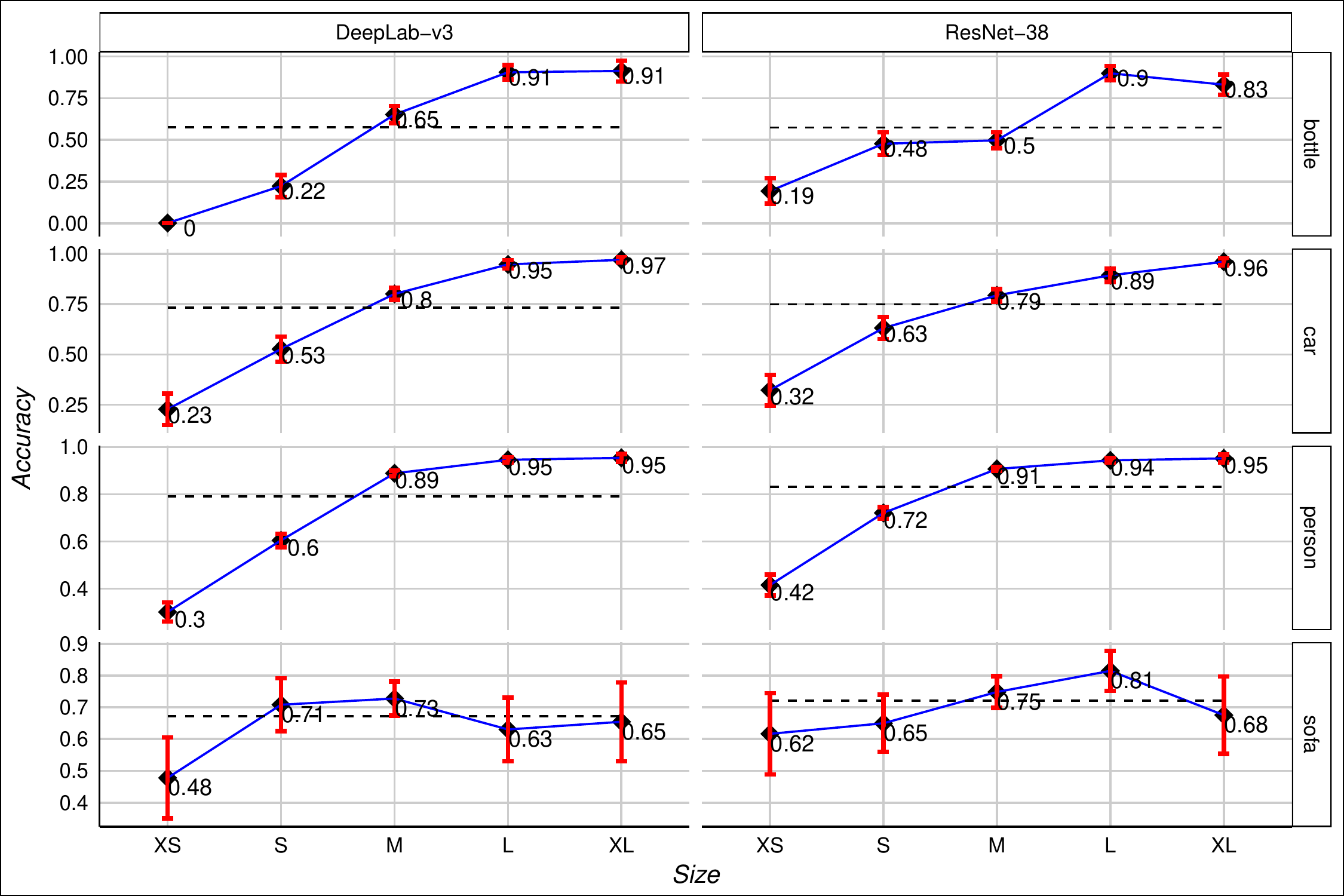}
		\end{minipage}%
	}
	\subfloat[CityScapes - Aspect Ratio\label{fig:cs-ar-size-sen}]{%
		\begin{minipage}{0.5\linewidth}
			\centering
			\includegraphics[width = 1.\linewidth, trim=4 20 4 4,clip]{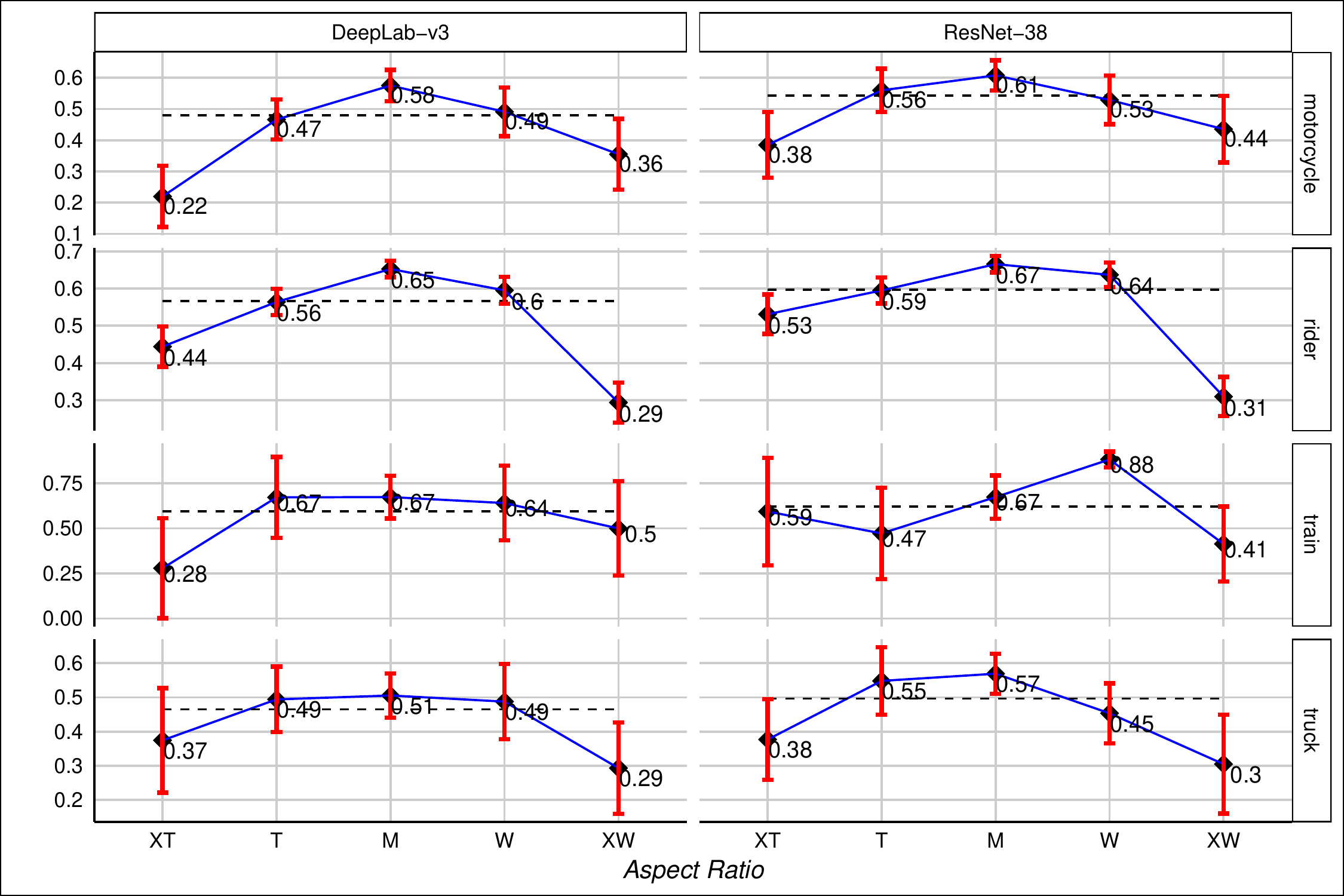}
		\end{minipage}%
	}
	\caption{Sensitivity of DeepLab-v3 and ResNet-38 to instance size on PASCAL VOC (a) and aspect ratio on CityScapes (b). Black diamond points with numbers represent average per-instance accuracy of given class and category; red lines indicate standard error bars, where black dashed lines denote average per-instance accuracy of the class (across all categories)
	}
	\label{fig:ar-size-sen}
	\vskip -0.15in
\end{figure}

Both models exhibit similar behaviour, under-performing on categories of small sizes and extreme aspect ratios, and preferring as larger instances as possible without significant variation in aspect ratio (Fig.~\ref{fig:ar-size-sen}). Nevertheless, there are some differences: e.g., DeepLab-v3 completely misses extra-small bottles and does steadily losses several points against ResNet-38 on all other classes with S or XS instances (except for small sofas) on VOC (Fig.~\ref{fig:voc-ar-size-sen}). Even though DeepLab-v3 steadily outperforms ResNet-38 across most categories in terms of globally computed metrics (Tables~\ref{table:res-voc} and~\ref{table:res-cs}), it does so mostly on average instances. Surprisingly, ResNet-38 seems to perform poorly on all trains except the wide ones (Fig.~\ref{fig:cs-ar-size-sen}).

\par
\begin{figure}[htb]
	\subfloat[PASCAL VOC\label{fig:voc-train}]{%
		\begin{minipage}{0.5\linewidth}
			\centering
			\includegraphics[width = 1.\linewidth, trim=4 2 4 4,clip]{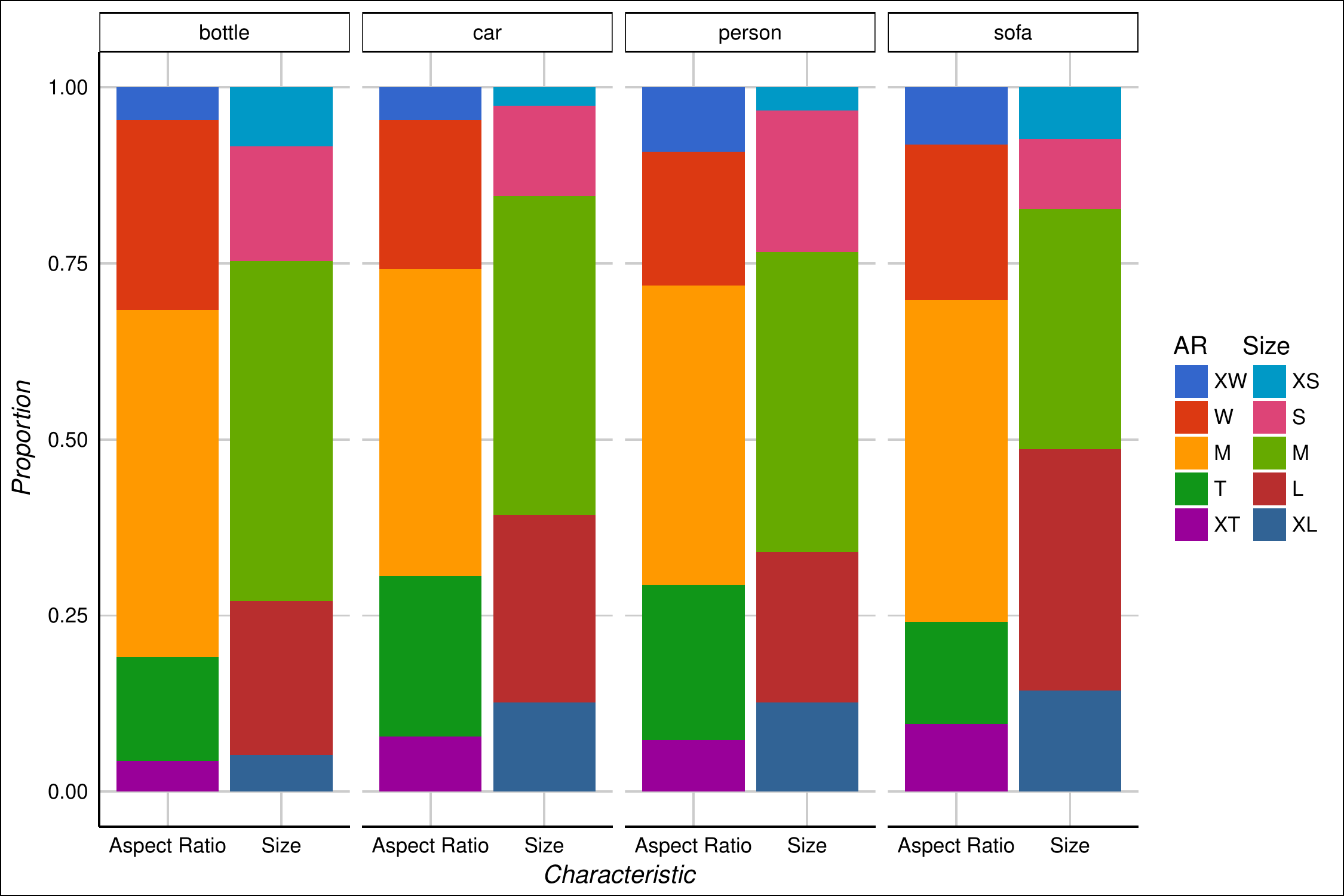}%
		\end{minipage}%
	}
	\subfloat[CityScapes\label{fig:cs-train}]{%
		\begin{minipage}{0.5\linewidth}
			\centering
			\includegraphics[width = 1.\linewidth, trim=4 2 4 4,clip]{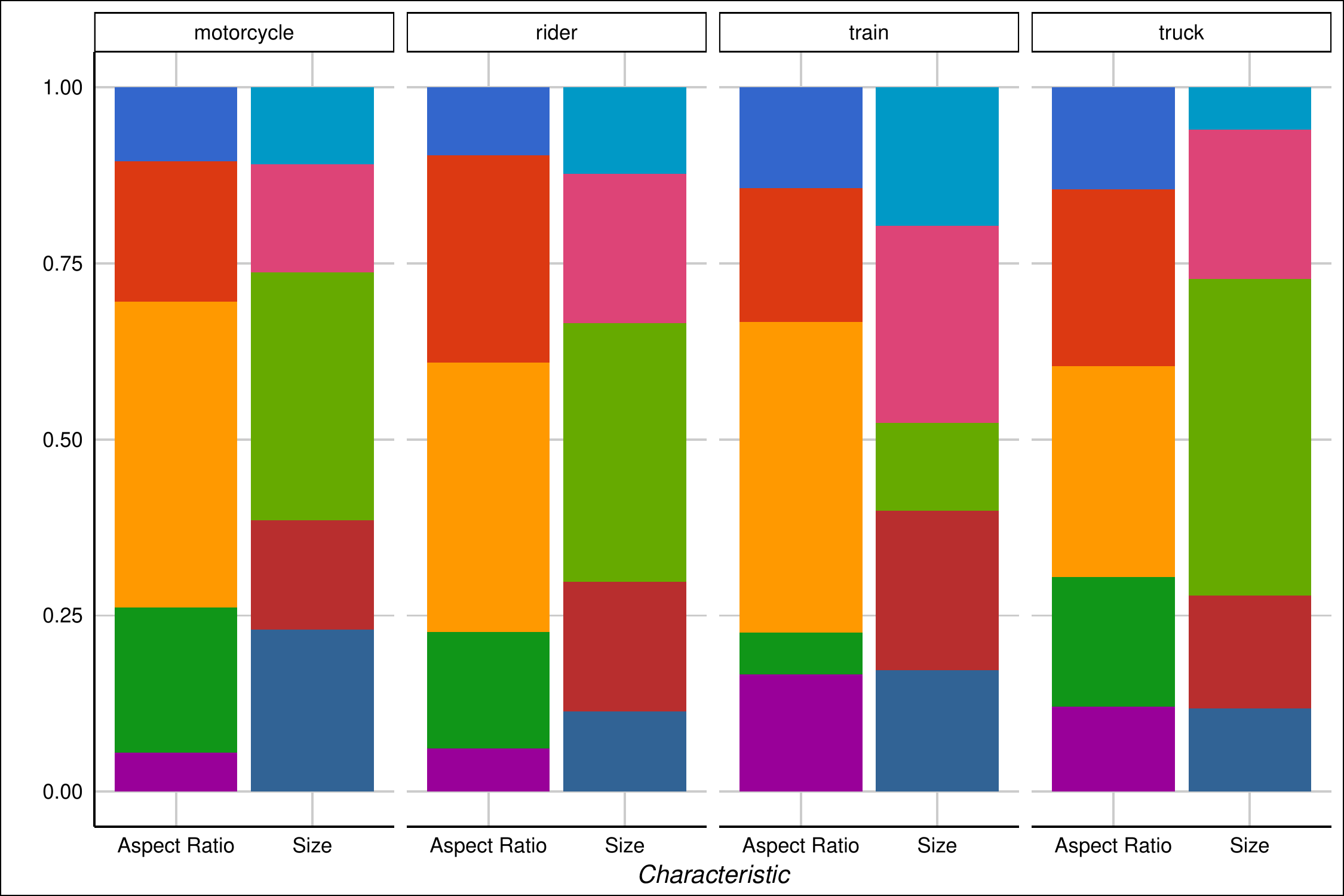}%
		\end{minipage}%
	}
	\caption{Distribution of instances of different sizes and aspect ratios on the training set of PASCAL VOC (a) and CityScapes (b)
}
\label{fig:train}
\end{figure}

Similar performance of two different models urges us to look at the distribution of different categories as present in the training set. As both of the models were using weights pre-trained from ImageNet~\cite{DengDSLL009}, and later fine-tuned for semantic segmentation on MS COCO~\cite{LinMBHPRDZ14} with possibly different choices of training data, we only consider the training set of PASCAL VOC augmented with annotations from BSD~\cite{HariharanABMM11} (in total, $10582$ images) in case of VOC, and the training set of CityScapes with $2975$ images. Comparing the performance results with the distribution across categories (Fig.~\ref{fig:train}), we note that there does seem to be the lack of extra-small instances (although not for the problematic class `bottle'), along with the shortage of instances with extreme aspect ratio (pointing to the similarity with the validation set).

\subsubsection{Extension}\label{subsub:ext-size} 
\begin{figure}[htb]
	\centering
	\begin{tabular}{ccccc}
		\raisebox{0.15\height}{\subfloat{\includegraphics[width = 0.24\linewidth]{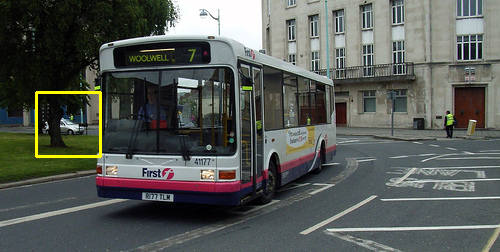}}} &
		\subfloat{\includegraphics[width = 0.18\linewidth]{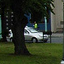}} &
		\subfloat{\includegraphics[width = 0.18\linewidth]{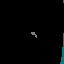}} &
		\subfloat{\includegraphics[width = 0.18\linewidth]{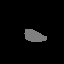}} &
		\subfloat{\includegraphics[width = 0.18\linewidth]{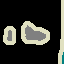}}\\[-0.15in]
		\subfloat{\includegraphics[width = 0.24\linewidth]{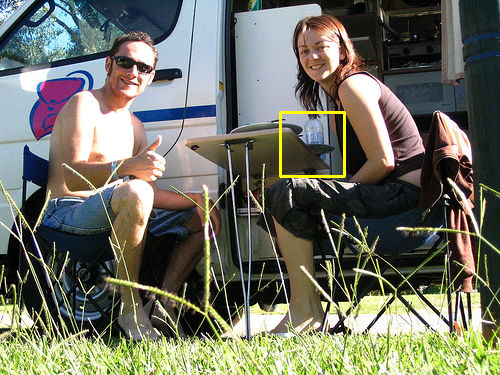}} &
		\subfloat{\includegraphics[width = 0.18\linewidth]{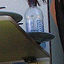}} &
		\subfloat{\includegraphics[width = 0.18\linewidth]{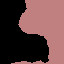}} &
		\subfloat{\includegraphics[width = 0.18\linewidth]{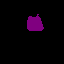}} &
		\subfloat{\includegraphics[width = 0.18\linewidth]{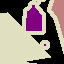}}\\[-0.15in]
		\subfloat{\includegraphics[width = 0.24\linewidth]{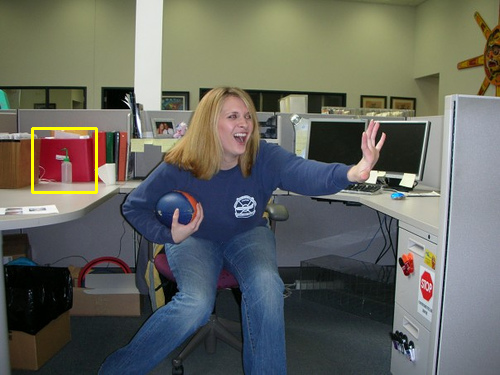}} &
		\subfloat{\includegraphics[width = 0.18\linewidth]{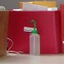}} &
		\subfloat{\includegraphics[width = 0.18\linewidth]{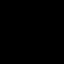}} &
		\subfloat{\includegraphics[width = 0.18\linewidth]{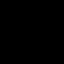}} &
		\subfloat{\includegraphics[width = 0.18\linewidth]{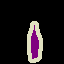}}\\
		Image&Crop&Orig Pred&Refined Pred&GT
	\end{tabular}
	\caption{Size experiments on classes `car' and `bottle' of VOC using DeepLab-v3. The yellow rectangle denotes the region that is resized and feed back into the network. Refined predictions are defined as the predictions from the second forward pass of the cropped region. White colour in the ground truth mask denotes `difficult' class, which is ignored during evaluation\label{fig:size-exp}}
\end{figure}

In an attempt to improve performance on instances of small and extra-small sizes, we conduct a simple experiment on classes `car' and `bottle' from VOC using DeepLab-v3. Concretely, we select all the images from the validation set that contain only $1$ instance of each of those classes belonging to either small (S) or extra-small (XS) size categories. In total, we found $19$ such images - $8$ with cars ($7$S and $1$XS) and $11$ with bottles ($7$S and $4$XS). We propagate each image through the DeepLab-v3 network, and for each we consider a score map corresponding to the class present (i.e., either `car' or `bottle'). Inside the score map we find the point with the highest activation score, and do a square crop (of size $64\times64$) around that point in the original image. We perform $4\times$ bicubic upsampling of the crop and propagate it through the network. Afterwards, we replace original predictions inside the cropped region with the new scores, and take the index of the highest class as the predicted label. This significantly improves the performance on small and extra small instances (Table~\ref{table:ext-size}), finding objects that were treated as background during the first forward pass (Fig.~\ref{fig:size-exp}).\par

\vskip -0.0in
\vskip 0.01in

Alternatively, to provide a baseline for our simple approach, we consider the case when there exists access to ground truth annotations. Under this scenario, we crop the image around bounding box corresponding to the ground truth segmentation enlarged by $16$ pixels in each direction, and upsample the crop $4\times$ using bicubic interpolation before feeding it into the network. As expected, this does further improve performance (Table~\ref{table:ext-size}), though it is very close to the one achieved without knowledge of ground truth. 

\vskip -0.3in
\setlength{\tabcolsep}{4pt}
\begin{table}[htb]
	\begin{center}
\caption{Accuracy and intersection-over-union on images of cars and bottles of small (S) and extra-small (XS) sizes from the validation set of PASCAL VOC using DeepLab-v3\label{table:ext-size}}
			\begin{tabular}{l|c|c|c|c|c}
				\specialrule{.15em}{0em}{0em} 
				 \textbf{Method}&\textbf{Class}& \multicolumn{2}{c|}{\textbf{Ins.-Wise Acc.,\%}} & \multicolumn{2}{|c}{\textbf{Total, \%}}\T\B\\
				\specialrule{.1em}{0em}{0em}
				&&XS&S&IoU&Acc\T\\
				\specialrule{.1em}{0em}{0em}
				\textbf{Orig}&Bottle&0.34&0.30&0.35&0.36\T\\
				&Car&0&42.48&40.96&40.96\T\\
				\specialrule{.1em}{0em}{0em}
				\textbf{Crop around max. act.}&Bottle&32.61&41.51&34.81&39.67\T\\
				&Car&0&81.48&65.81&75.06\T\\
				\specialrule{.1em}{0em}{0em}
				\textbf{Crop around GT bbox}&Bottle&52.76&56.63&51.28&57.20\T\\
				&Car&11.63&81.44&70.11&75.26\B\\
				\specialrule{.15em}{0em}{0em}
			\end{tabular}
\end{center}
\vskip -0.5in
\end{table}

\subsubsection{Recommendations}
Failure to recognise objects with extreme characteristics is prevalent across different domains, including object detection~\cite{HoiemCD12,HariharanAGM14}. As shown by Li~\etal~\cite{LiLWXFY17}, this is often due to feature mismatch between small and large instances. To alleviate such an imbalance, they trained a generative adversarial network so that the network would mimic the features of large objects on the small ones, effectively fooling the detector to recognise the small object as if it was large. The extension of this approach can easily be adapted for semantic segmentation. Additionally, as shown above, attention-based post-processing transformations may reduce the need to train a separate model to deal with peculiar object instances, and can be extremely helpful.

\subsection{Error Taxonomy}
Here we present the results for classes `bicycle', `chair', `dog' and `sofa' (VOC), and `bus', `motorcycle', `road' and `traffic light' (CityScapes), as these classes demonstrate well connections between model performance and different types of errors among all the classes.
\vskip -0.15in
\vskip -5in
\subsubsection{Observations}
\begin{figure}[htb]
	\subfloat[PASCAL VOC\label{fig:voc-prop-err}]{%
		\begin{minipage}{0.5\linewidth}
			\centering
			\includegraphics[width = 1.\linewidth, trim=4 2 4 4,clip]{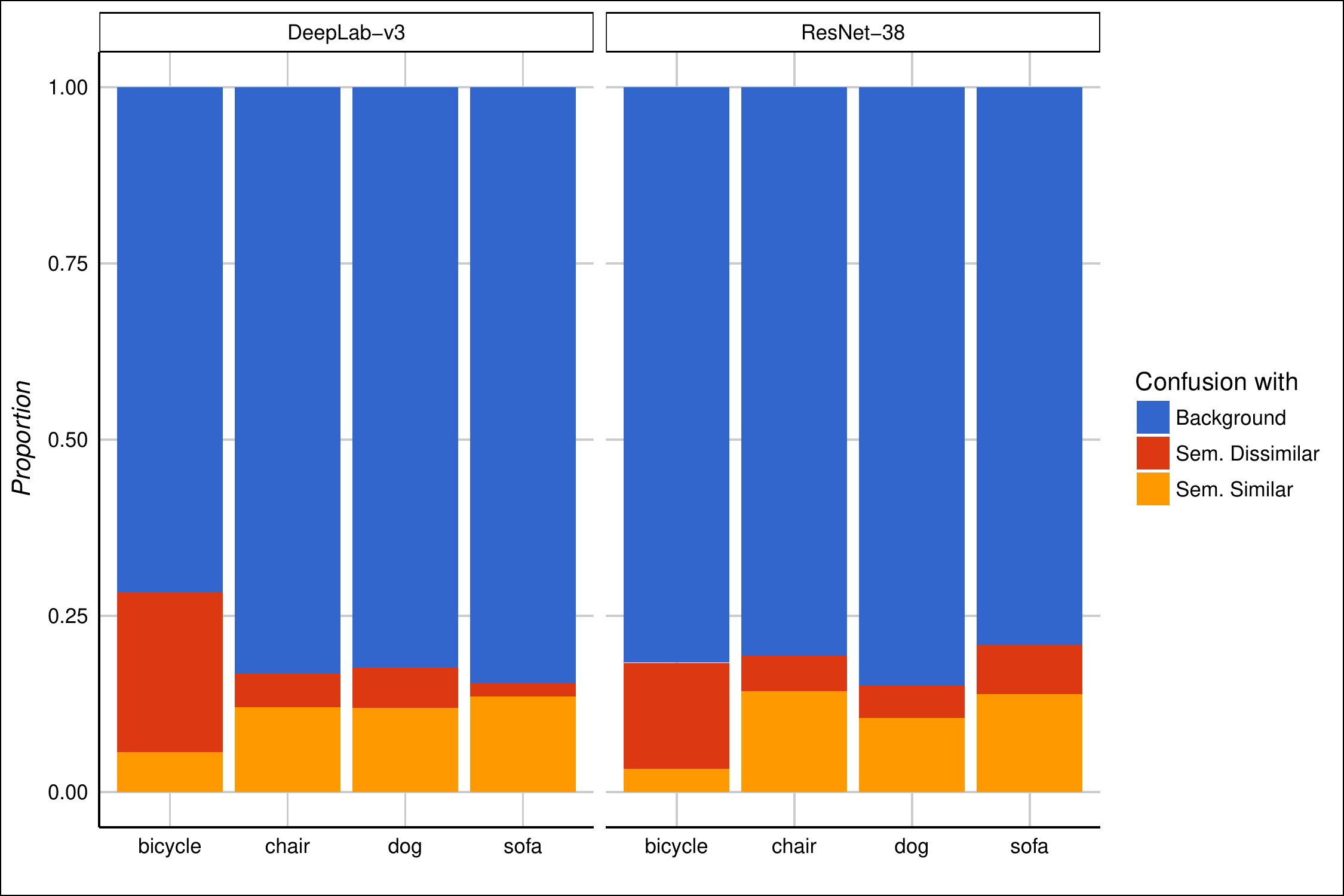}%
		\end{minipage}%
	}
	\subfloat[CityScapes\label{fig:cs-prop-err}]{%
		\begin{minipage}{0.5\linewidth}
			\centering
			\includegraphics[width = 1.\linewidth, trim=4 2 4 4,clip]{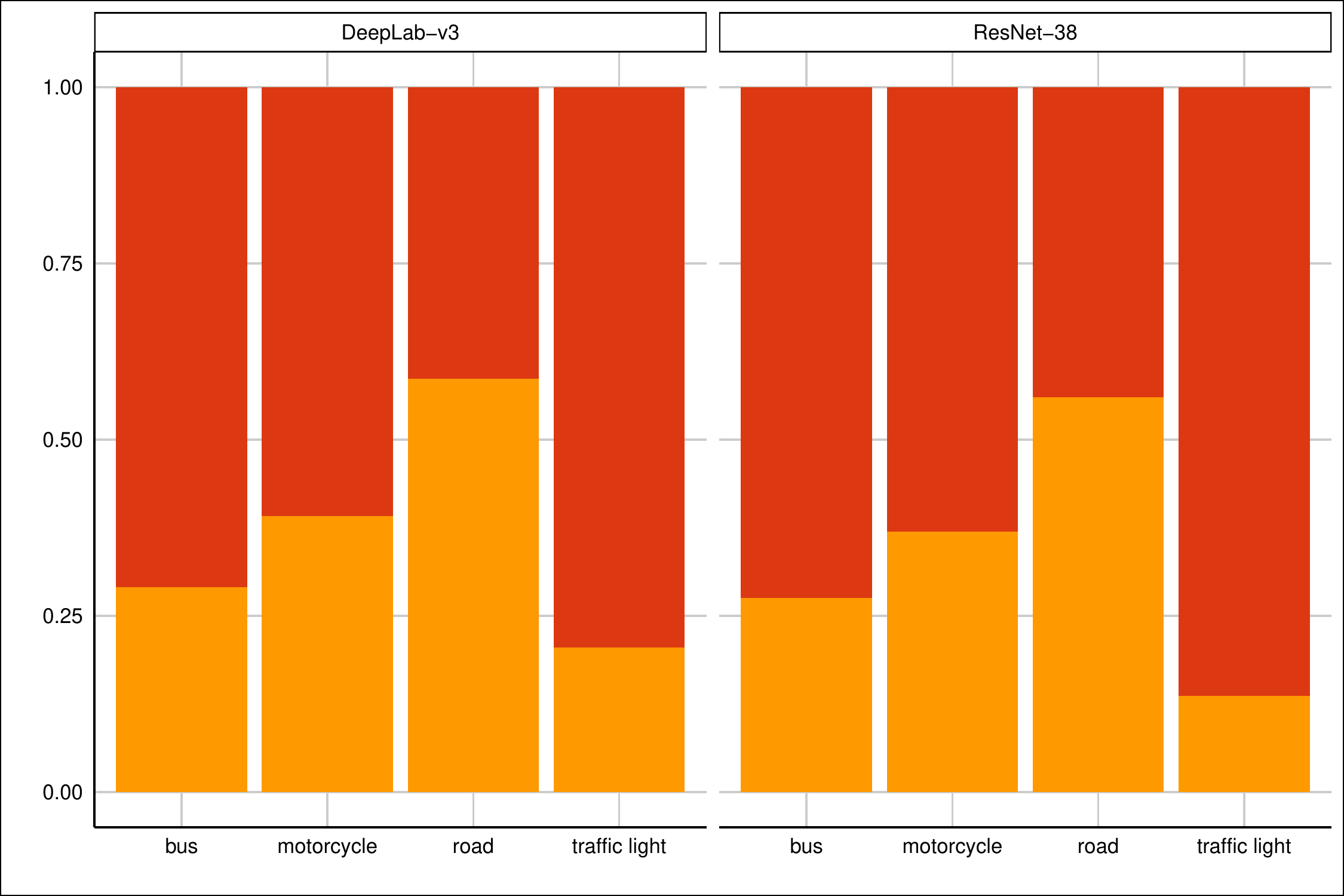}%
		\end{minipage}%
	}
	\caption{Proportion of errors caused by confusion with other classes on the validation set of PASCAL VOC (a) and CityScapes (b)
	}
\label{fig:prop-err}
\vskip -0.35in
\end{figure}

\begin{figure}[htb]
	\subfloat[PASCAL VOC\label{fig:voc-sem-sim}]{%
		\begin{minipage}{0.5\linewidth}
			\centering
			\includegraphics[width = 1.\linewidth, trim=4 20 4 4,clip]{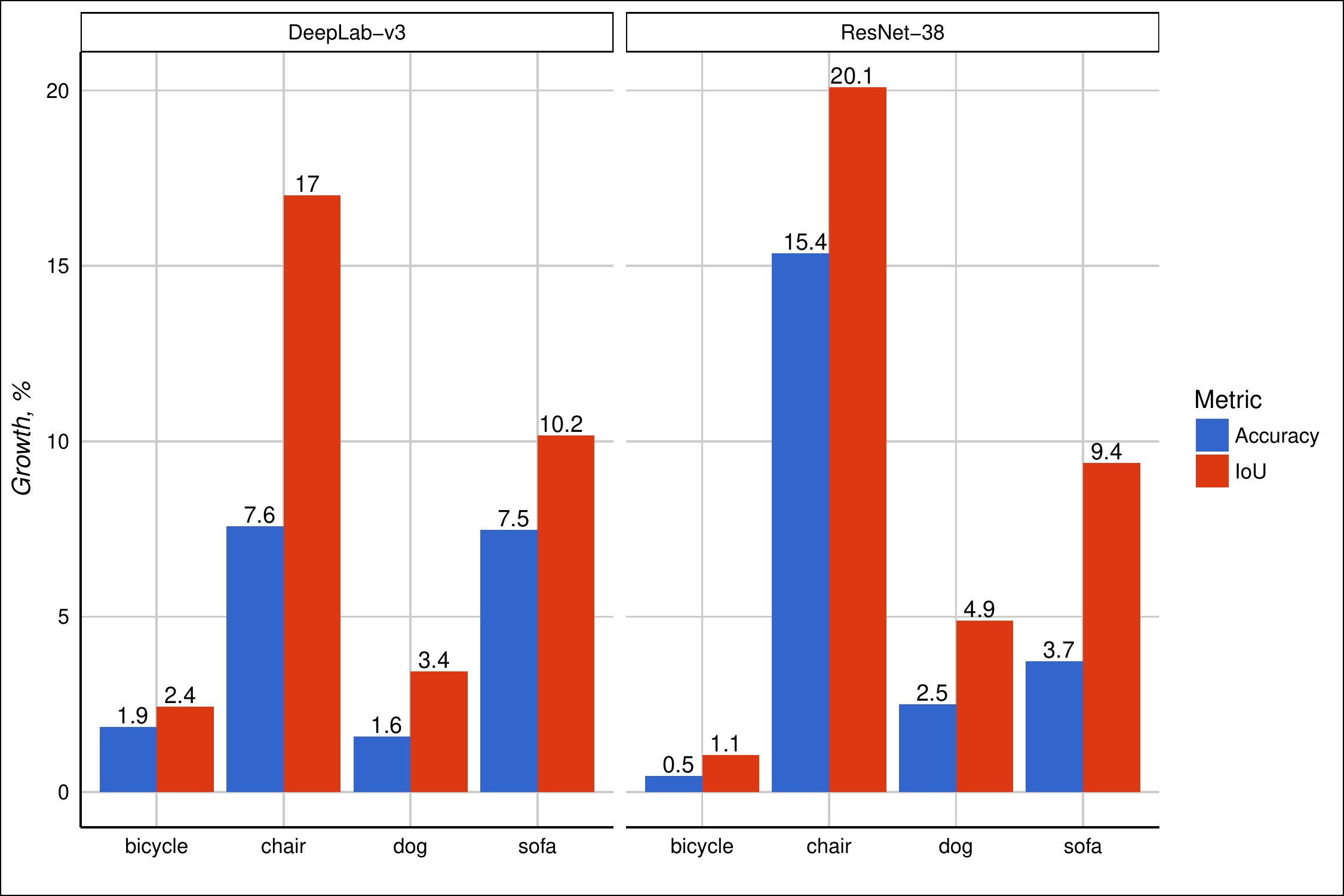}
		\end{minipage}%
	}
	\subfloat[CityScapes\label{fig:cs-sem-sim}]{%
		\begin{minipage}{0.5\linewidth}
			\centering
			\includegraphics[width = 1.\linewidth, trim=4 20 4 4,clip]{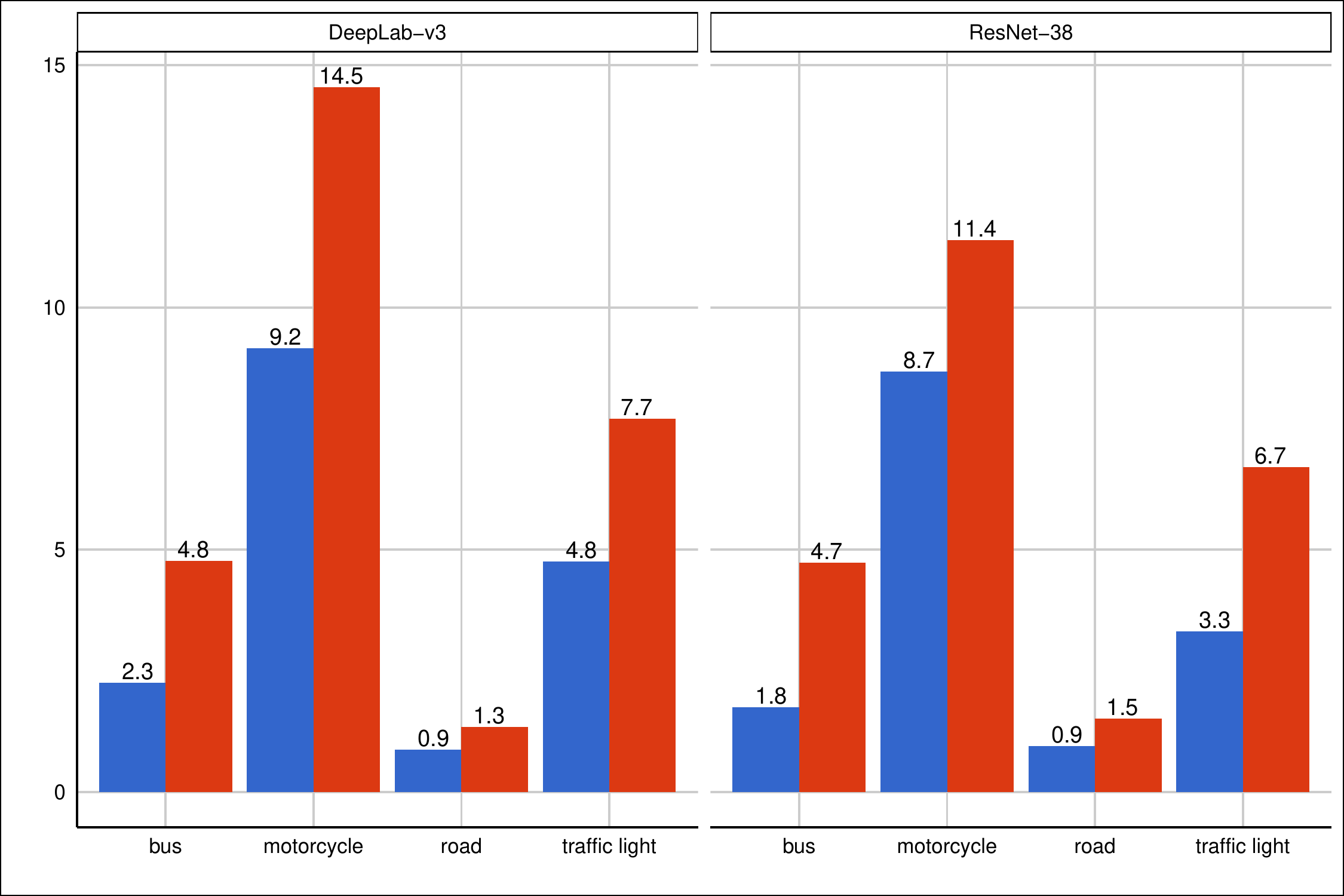}
		\end{minipage}%
	}
	\caption{Accuracy and IoU growth on the validation sets of PASCAL VOC (a) and CityScapes (b) if semantically similar labels were to be considered as one\label{fig:sem-sim}
	}
\vskip -0.2in
\end{figure}

If the model is confused, it is more likely to be confusion with background, if such a class exists, or with semantically dissimilar classes~(Fig.~\ref{fig:prop-err}). 
Class `bicycle` is rarely confused with other vehicles, while `chair' and `sofa' are often confused with each other~(Fig.~\ref{fig:voc-prop-err}). For CityScapes, the situation is similar, although not for `road', where the proportion of errors is practically equal~(Fig.~\ref{fig:cs-prop-err}). In some applications, one can treat predictions of semantically similar classes as belonging to a single class, and if we were to follow this approach, we would witness significant gains across the chosen classes (Fig.~\ref{fig:sem-sim}). As evident from Fig.~\ref{fig:cs-sem-sim}, class `road' experiences a smaller performance improvement even with a large proportion of errors caused by confusion with semantically similar classes. This happens as performance on this class is already high - almost $99\%$ (Table~\ref{table:res-cs}), and the number of errors to be corrected is small.\par
\begin{figure}[htb]
	\subfloat[PASCAL VOC - Accuracy\label{fig:voc-misloc}]{%
		\begin{minipage}{0.5\linewidth}
			\centering
			\includegraphics[width = 1.\linewidth, trim=4 2 1 4,clip]{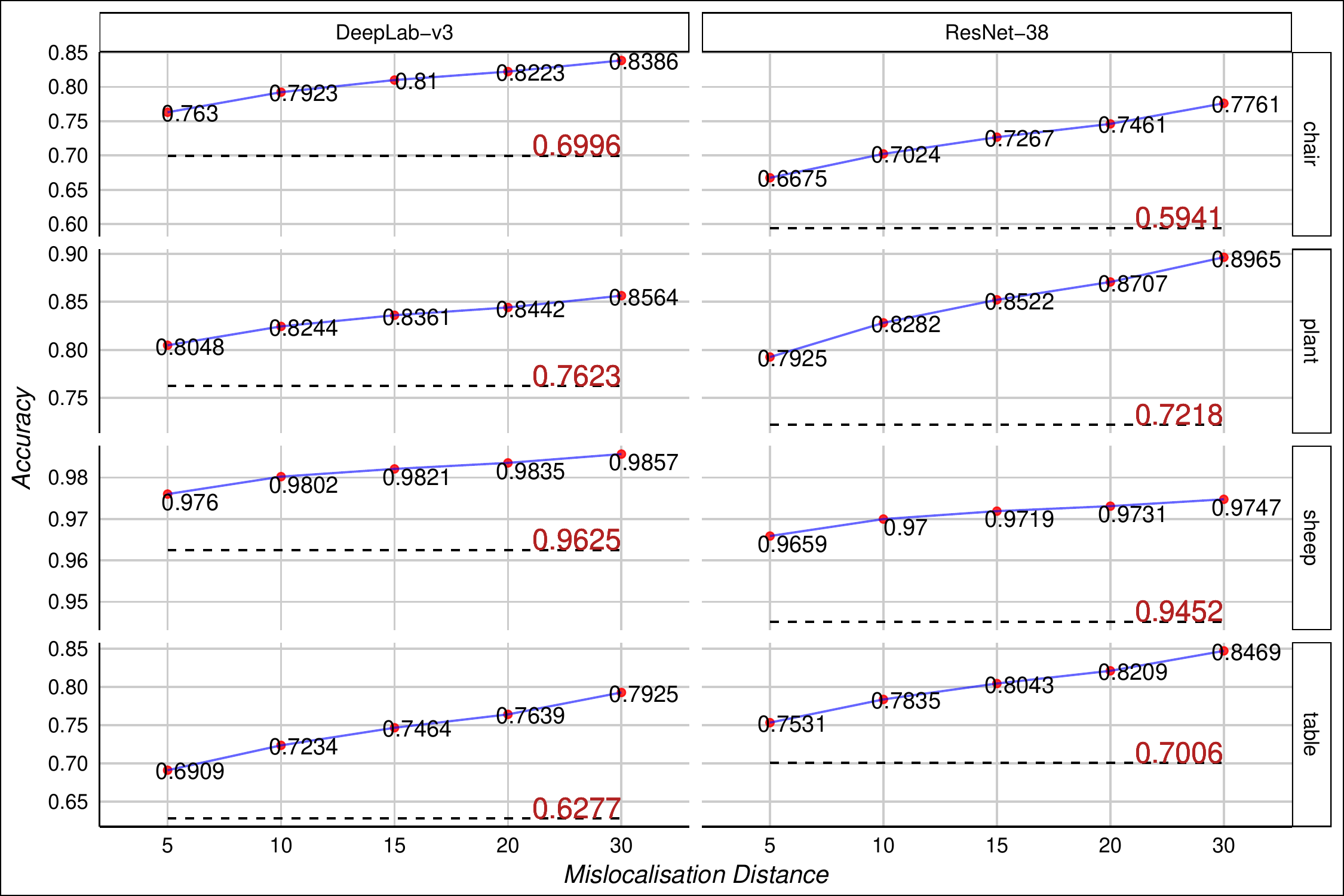}%
		\end{minipage}%
	}
	\subfloat[CityScapes - IoU\label{fig:cs-misloc}]{%
		\begin{minipage}{0.5\linewidth}
			\centering
			\includegraphics[width = 1.\linewidth, trim=4 2 1 4,clip]{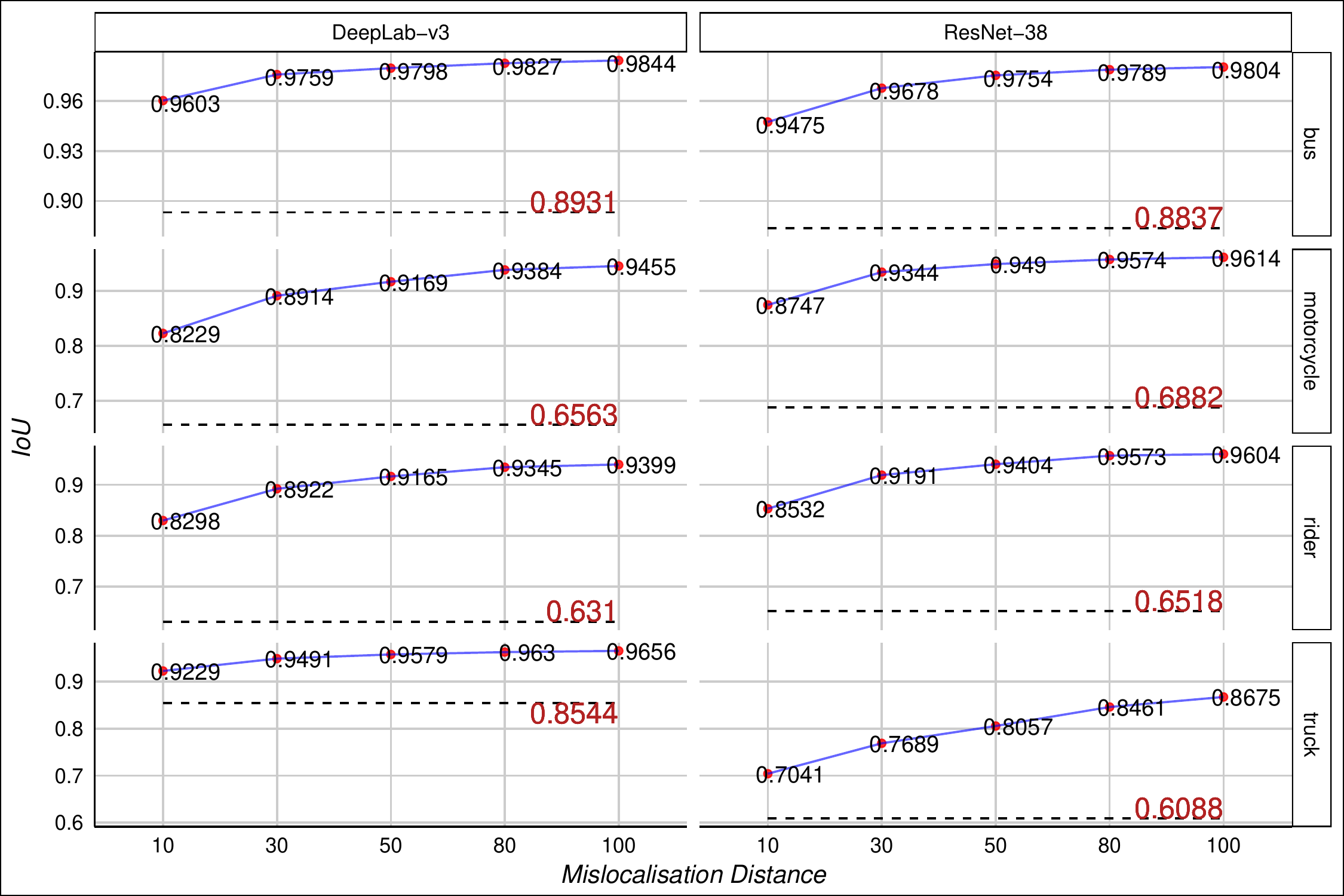}%
		\end{minipage}%
	}
	\caption{Gains in accuracy on the validation set of PASCAL VOC (a) and intersection-over-union on the validation set of CityScapes (b) if mislocalisation errors were to be corrected. Red points with numbers represent average metrics of a given class for a certain mislocalisation distance (half-length of the square patch centered at the point of prediction); black dashed lines with numbers denote mean average metric of the class (without any error correction)
	}
\label{fig:misloc}
\vskip -0.15in
\end{figure}

We further look at how mislocalisation errors affect the models. For VOC, we consider square crops centered at the point of prediction with half-side lengths being equal to $5$, $10$, $15$, $20$ and $30$ pixels; for CityScapes that contains high-resolution images those values are $10$, $20$, $50$, $80$ and $100$ pixels, respectively. If the predicted label does exist inside the square crop of the ground truth map, then the prediction is considered to be correct. From Fig.~\ref{fig:misloc}, we can easily notice that for all the chosen classes across the datasets even the window with the smallest size gives rise to a significant boost - from around $5\%$ to even more than $20\%$, for both accuracy and IoU. This is tangible, and in situations where time for additional post-processing might be available, it can be exploited to correct mislocalisation errors, for example, by analogy to a simple zoom-in strategy outlined in Sect.~\ref{subsub:ext-size}.

\vskip -0.15in

\subsubsection{Extension}
Apart from choosing to sacrifice any differentiation between semantically similar classes, or to lose exact per-pixel classification, we propose another straightforward approach able to eliminate both types of errors. In particular, we consider the effect of selecting top-$N$ predictions at each pixel and looking at whether any of top-$N$ predicted labels is the ground truth one. For practical applications, this can make a huge difference, effectively reducing the search space to only few labels and treating additional post-processing steps as simply binary or ternary classification. For example, in medical imaging, a segmentation model might be initially solving the labelling problem with more than $10$ classes and solely relying on one class prediction with somewhat unstable performance, might not be the best strategy available. In contrast, as evident from Table~\ref{table:ext-top-n}, even considering top-$2$ classes with highest scores does significantly boost the numbers, possibly leading to better performance with auxiliary post-processing.

\begin{table}[htb]
	\begin{center}
		\caption{Top-N results on the validation set of CityScapes. Predictions for which any of top-N scores corresponds to the ground truth label are deemed correct
			\label{table:ext-top-n}}
		\begin{tabular}{c|c|c|c|c|c|c}
			\specialrule{.15em}{0em}{0em} 
			Metric & Model & Top-1 & Top-2 & Top-3 & Top-4 & Top-5\T\B\\
			\specialrule{.1em}{0em}{0em}
			\hline
			\parbox[t]{1mm}{\multirow{2}{*}{\rotatebox[origin=c]{90}{mAcc}}} & DeepLab-v3 & 86.38 & 94.76 & 97.33 & 98.41 & 98.95\T\\
			& ResNet-38 & 84.85 & 94.36 & 97.33 & 98.56 & 99.10\B\\
			\specialrule{.15em}{0em}{0em}
			\hline
			\parbox[t]{1mm}{\multirow{2}{*}{\rotatebox[origin=c]{90}{mIoU}}} & DeepLab-v3 & 79.18 & 92.27 & 96.23 & 97.83 & 98.62\T\\
			& ResNet-38 & 77.67 & 91.94 & 96.29 & 98.03 & 98.78\B\\
			\specialrule{.15em}{0em}{0em}
		\end{tabular}
	\end{center}
	\vskip -0.25in
\end{table}
\setlength{\tabcolsep}{1.4pt}

\subsubsection{Recommendations}
Confusion with background and with dissimilar classes tends to occupy the largest portion of errors committed by the model. To eliminate the effect of background, one might consider to divide it into more classes, effectively providing more information for the model to learn from. E.g., Hu~\etal~\cite{abs-1711-10370} proposed to exploit a transfer function to acquire semantic segmentation for $3000$ classes having only bounding box annotations. Furthermore, considering structured loss functions can help in alleviating the effect of the errors~\cite{Berman_2018_CVPR}.

\subsection{Uncertainty}
First of all we note that for the uncertainty experiments we make comparisons within the model itself, not between the models, as the scale factor of logits in all the models is different causing the softmax probabilities and, consequently, relative entropy and relative probability to be different, as well.  We present the results for classes `bird', `cat', `chair', `sheep' (VOC), and `bicycle', `bus', `person', `train' (CityScapes), as these classes showcase diverse patterns prevalent among all the classes. 


\begin{figure}[htb]
	\subfloat[PASCAL - rel. entropy - object size\label{fig:voc-ent-size}]{%
		\begin{minipage}{0.5\linewidth}
			\centering
			\includegraphics[width = 1.\linewidth, trim=4 20 4 4,clip]{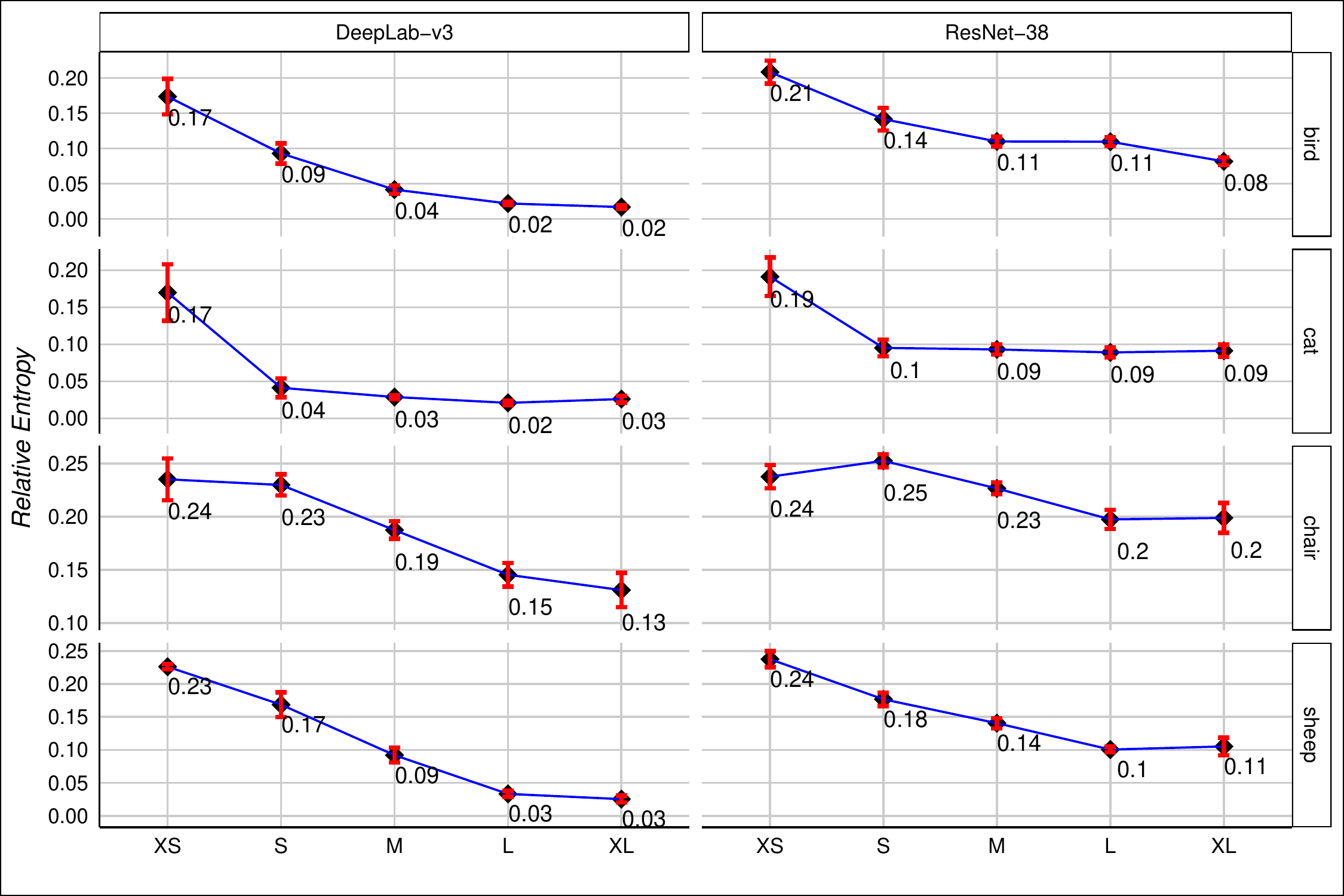}
		\end{minipage}%
	}
	\subfloat[CityScapes - rel. probability - asp. ratio\label{fig:cs-pbdiff-ar}]{%
		\begin{minipage}{0.5\linewidth}
			\centering
			\includegraphics[width = 1.\linewidth, trim=4 20 4 4,clip]{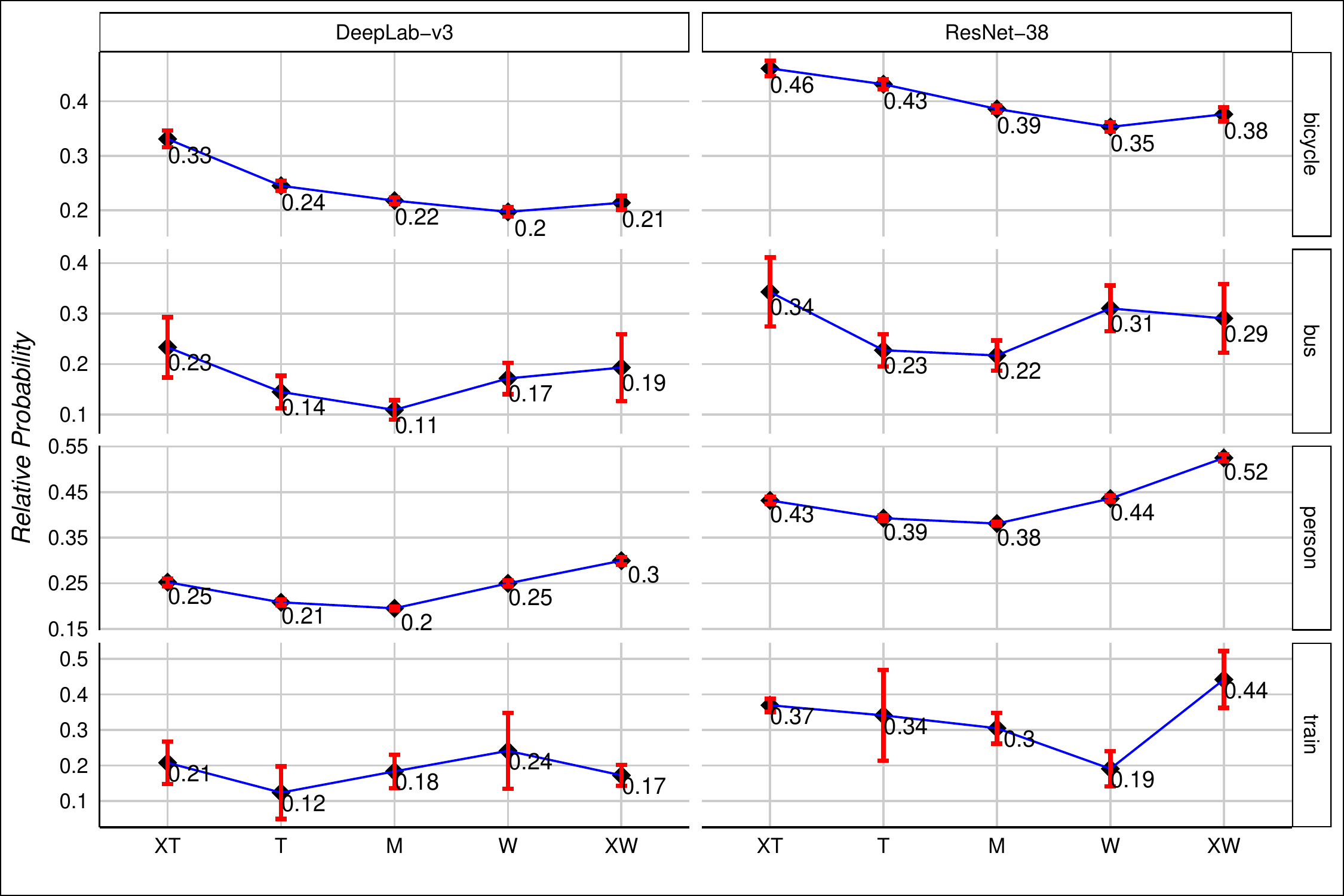}
		\end{minipage}%
	}
	\caption{Relative entropy on PASCAL VOC as function of instance size (a) and relative probability on CityScapes as function of aspect ratio (b)}
	\label{fig:unc-obj}
\vskip -0.15in
\end{figure}

\begin{figure}[htb]
	\subfloat[PASCAL VOC - rel. entropy\label{fig:voc-ent-dist}]{%
		\begin{minipage}{0.5\linewidth}
			\centering
			\includegraphics[width = 1.\linewidth, trim=4 20 4 4,clip]{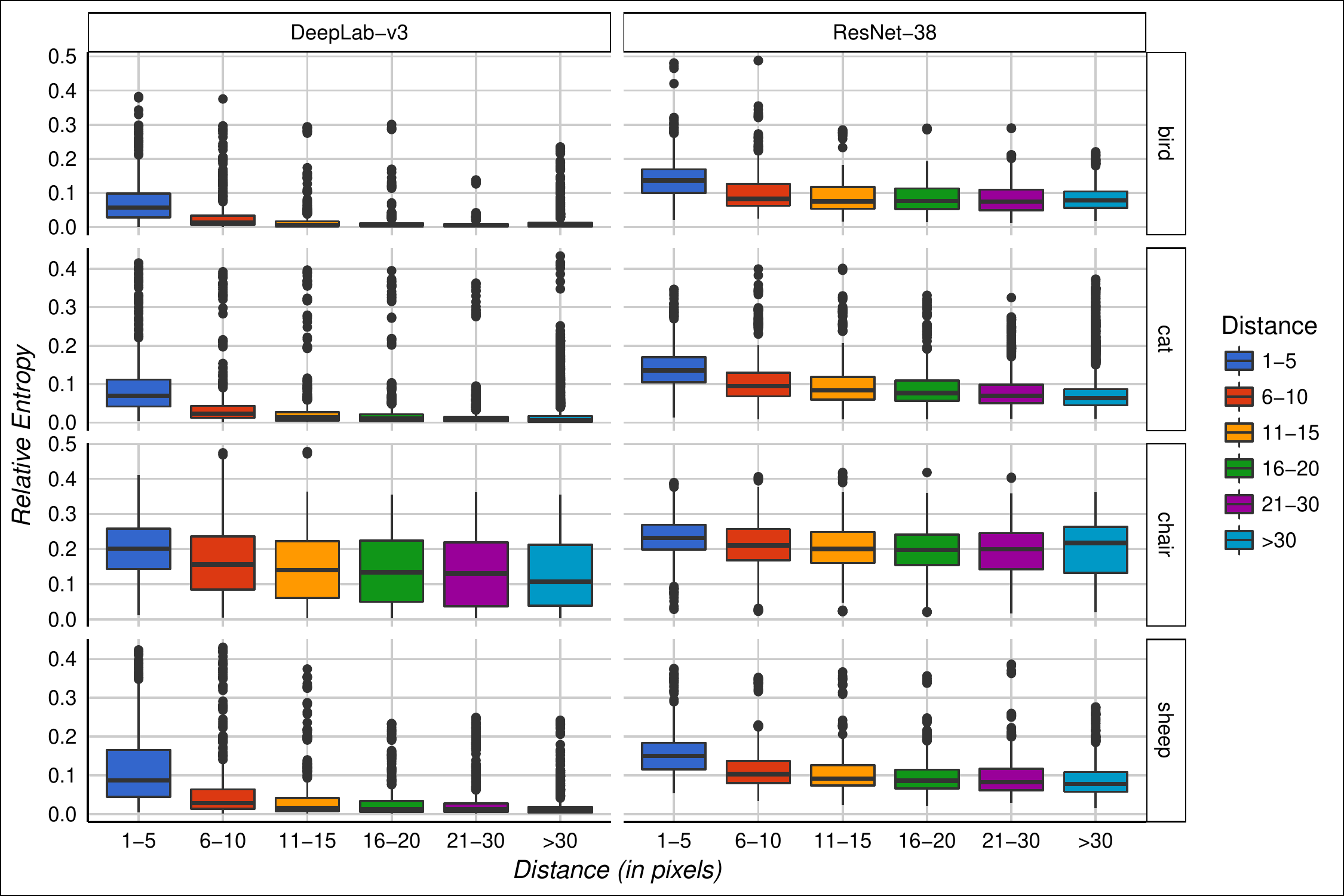}
		\end{minipage}%
	}
	\subfloat[CityScapes - rel. probability\label{fig:cs-pbdiff-dist}]{%
		\begin{minipage}{0.5\linewidth}
			\centering
			\includegraphics[width = 1.\linewidth, trim=4 20 4 4,clip]{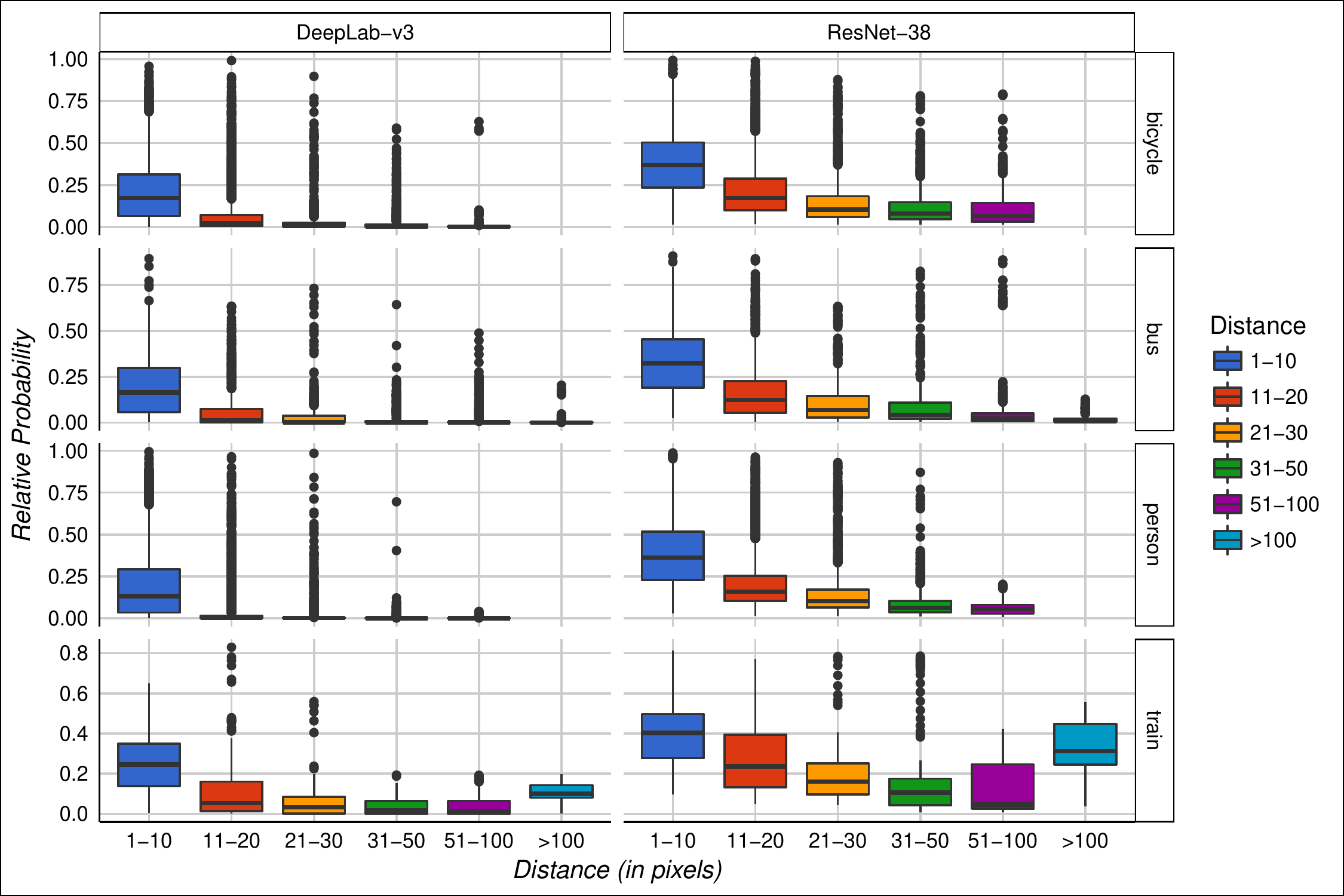}
		\end{minipage}%
	}
	\caption{Relative entropy on the validation set of PASCAL VOC (a) and relative probability on the validation set of CityScapes (b) as function of distance to the boundary (in pixels). The lower line of the box denotes the lower quartile ($25\%$), the black line inside the box depicts the median value, and the upper line of the box shows the upper quartile ($75\%$)}
	\label{fig:unc-dist}
\vskip -0.2in
\end{figure}

\begin{figure}[!htbp]
	\subfloat[PASCAL VOC - rel. probability\label{fig:voc-pbdiff-err}]{%
		\begin{minipage}{0.5\linewidth}
			\centering
			\includegraphics[width = 1.\linewidth, trim=4 20 4 4,clip]{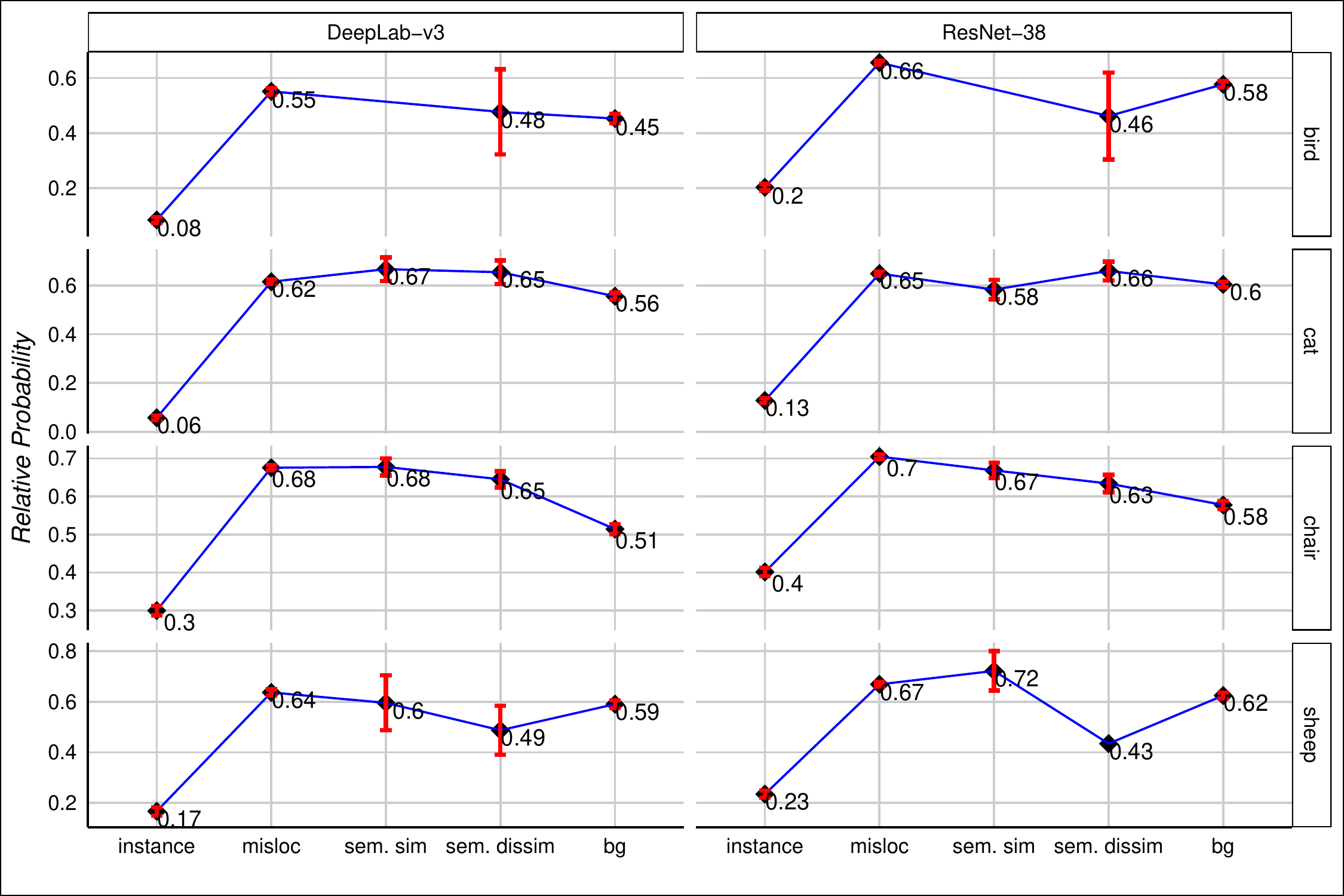}
		\end{minipage}%
	}
	\subfloat[CityScapes - rel. entropy\label{fig:cs-ent-err}]{%
		\begin{minipage}{0.5\linewidth}
			\centering
			\includegraphics[width = 1.\linewidth, trim=4 20 4 4,clip]{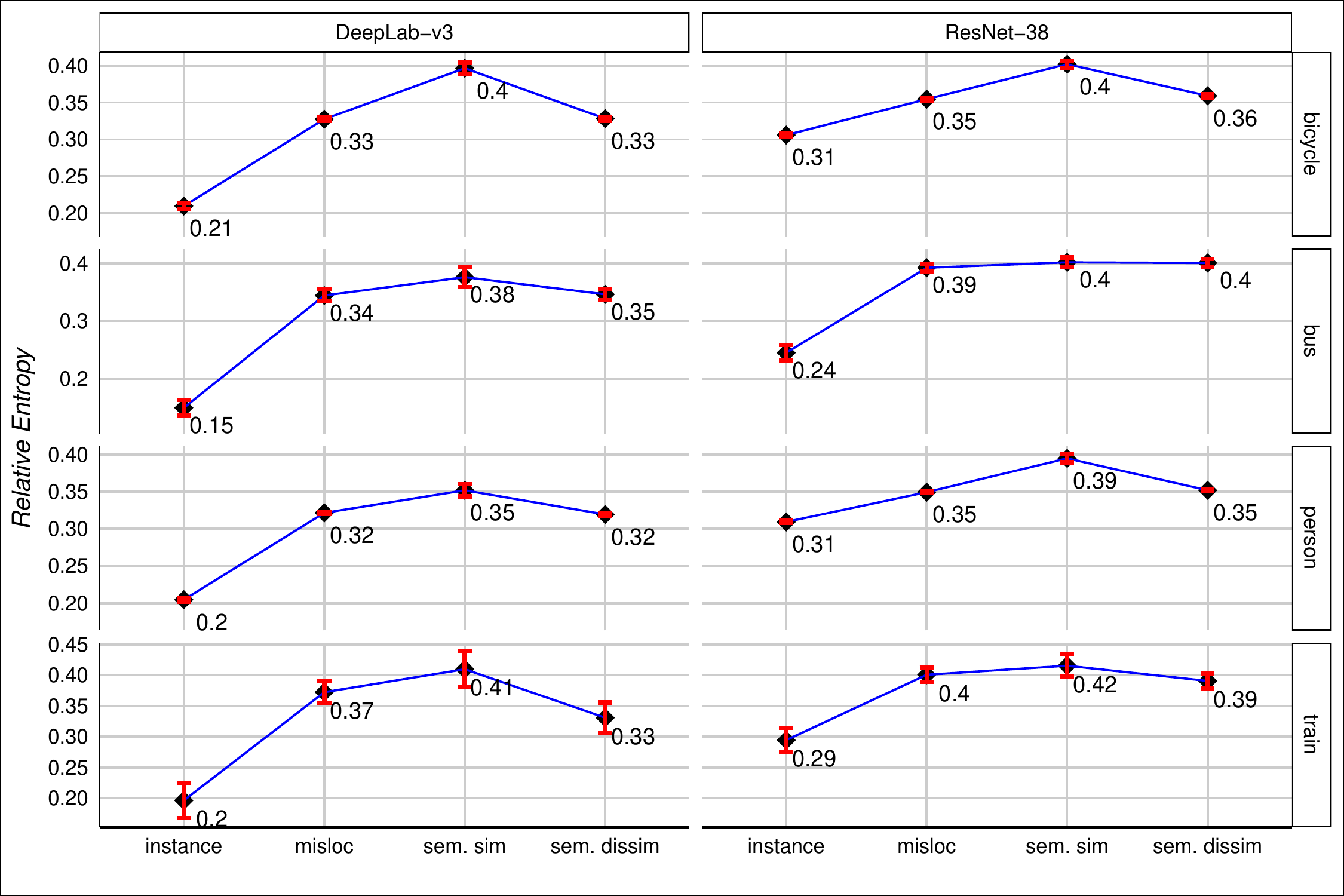}
		\end{minipage}%
	}
	\caption{Relative probability on the validation set of PASCAL VOC  (a) and relative entropy on the validation set of CityScapes (b) as function of instance-wise uncertainty and different types of errors. \textit{Instance} stands for average uncertainty per-instance, \textit{misloc} stands for average uncertainty at points with mislocalisation errors, and \textit{sem. sim.}, \textit{sem. dissim.} and \textit{bg} shows average uncertainty at points with confusion errors, caused by confusion with either semantically similar, or dissimilar classes, or with background, respectively}
	\label{fig:err}
\vskip -0.2in
\end{figure}

\subsubsection{Observations}
We first look at how the defined notions of uncertainty behave on instances of different sizes and aspect ratios (Fig.~\ref{fig:unc-obj}). The models tend to be less certain about smaller objects with extreme aspect ratio, which is inversely proportional to the behaviour of accuracy against object characteristics (Fig.~\ref{fig:ar-size-sen}). The tendency to be more certain on average about larger objects can be explained by the distance to the boundary from the point of prediction - for larger objects, this distance tends to be bigger, which, in turn, make uncertainty smaller (Fig.~\ref{fig:unc-dist}).\par
Additionally, we consider how uncertainty differs across the range of different types of errors, as well as the average uncertainty per instance. As evident from Fig.~\ref{fig:err}, even when the models commit errors, their uncertainty might signal about the error, which may be helpful in lots of scenarios. In particular, relative probability tends to be the highest at the points with mislocalisation errors (of radius $5$ for PASCAL VOC and $10$ for CityScapes), followed by confusion with semantically similar classes~(Fig.~\ref{fig:voc-pbdiff-err}), signalling that top-$2$ scores are practically equal. In contrast, relative entropy seems to be the highest on semantically similar classes closely followed by mislocalisation errors (Fig.~\ref{fig:cs-ent-err}), indicating that there is no clear winner class amongst predictions.

\subsubsection{Extension}
As uncertainty seems to take higher values when the model commits an error, here we take a closer look at the ability of uncertainty to differentiate between foreground and background on PASCAL VOC. To this end, we consider relative entropy and relative probability computed on images from the validation set using ResNet-38. We treat pixels with uncertainty higher than the image average as `foreground' pixels, and all others as `background'. We compare the resultant masks against ground truth segmentations, and demonstrate our results in Table~\ref{table:ext-unc}. Both uncertainty based foreground-background predictors achieve solid accuracy, but the method using relative entropy suffers from a large number of false positives, as evident from precision, while the one with relative probability has a lower recall, signalling about a large number of undetected foreground pixels. Our simplistic way of thresholding is, of course, a subject of further improvements.

\setlength{\tabcolsep}{4pt}
\begin{table}[htb]
	\vskip -0.15in
	\begin{center}
		\caption{Foreground-background segmentation on the validation set of PASCAL VOC using different uncertainty measures
			\label{table:ext-unc}}
		\begin{tabular}{c|c|c|c}
			\specialrule{.15em}{0em}{0em} 
			Uncertainty & Precision,\% & Recall,\% & Accuracy,\%\T\B\\
			\specialrule{.1em}{0em}{0em}
			\hline
			Rel. Entropy & 30.16 & 47.35 & 56.69\T\\
			Rel. Probability & 42.23 & 43.72 & 69.03\B\\
			\specialrule{.15em}{0em}{0em}
		\end{tabular}

	\end{center}
	\vskip -0.45in
\end{table}
\setlength{\tabcolsep}{1.4pt}

\subsubsection{Recommendations}
Exploiting uncertainty is becoming a topic of its own in deep learning~\cite{KendallG17}, and we encourage practitioners and researchers to be aware of it. For semantic segmentation, additional post-processing techniques based on uncertainty measures may alleviate certain types of errors and might signal about a missing object, or even about a new unseen class. 

%


\vskip -0.15in
\section{Discussion \& Conclusions}

In this paper, we approached the question of diagnostics in semantic segmentation. This is an extremely broad area of research, and we believe that for further advances in the field we will need to get a better grasp on the current advances that we have. To this end, we laid out the extensive (but by no means the exclusive) categorisation of most prevalent sources of errors in semantic segmentation, along with novel points considering two types of uncertainty, as well as simple extensions. Our findings signal that the performance of semantic segmentation models has indeed reached high levels, and future advances should be concerned with how to unite different types of annotated data instead of pursuing expensive per-pixel labellings; how to exacerbate the effect of particular error sources; and how to make use of uncertainty to improve the stability of the model. We hope that this report will provide inspiration for a broader research into the question of how exactly segmentation models achieve such extraordinary results, as well as will bring more advances into the area.

Besides the above findings, we believe that an efficient usage of the models that we have now (i.e., transfer learning) must be explored further along with the notions of uncertainty for learning new objects and classes. We aim to pursue and address those directions in our future research.

\bibliographystyle{splncs04}
\bibliography{egbib}

\end{document}